\documentclass[letterpaper]{article} 
\usepackage{aaai25}  
\usepackage{times}  
\usepackage{helvet}  
\usepackage{courier}  
\usepackage[hyphens]{url}  
\usepackage{graphicx} 
\usepackage{natbib}  
\usepackage{caption} 
\usepackage{algorithm}
\usepackage{algpseudocode}

\usepackage{amssymb}

\usepackage{amsmath}

\usepackage{booktabs}
\usepackage{multirow}
\usepackage{xcolor}
\usepackage{cleveref}
\usepackage{mathrsfs}

\usepackage{amsmath}
\usepackage{amssymb}
\usepackage{mathtools}
\usepackage{amsthm}

\usepackage{algorithm}
\usepackage{algpseudocode}
\usepackage{cleveref}
\usepackage{multirow}

\theoremstyle{plain}

\theoremstyle{definition}

\theoremstyle{remark}

\usepackage[algo2e,ruled,linesnumbered]{algorithm2e}
\usepackage{rotating}
\usepackage{pdflscape}

\usepackage[textsize=tiny]{todonotes}

\usepackage{float}
\usepackage{subcaption}

\usepackage[textsize=tiny]{todonotes}

\usepackage{newfloat}
\usepackage{listings}

\DeclareCaptionStyle{ruled}{labelfont=normalfont,labelsep=colon,strut=off} 
\floatstyle{ruled}
\newfloat{listing}{tb}{lst}{}
\floatname{listing}{Listing}
\title{MetaSymNet: A Tree-like Symbol Network with Adaptive Architecture and Activation Functions}
\author{
    Yanjie Li\textsuperscript{\rm 1,3,4},
    Weijun Li\textsuperscript{\rm 1,2,3,4}\thanks{Corresponding author.},
    Lina Yu\textsuperscript{\rm 1}$^{*}$,
    Min Wu\textsuperscript{\rm 1}$^{*}$,
    Jingyi Liu\textsuperscript{\rm 1},
    Shu Wei\textsuperscript{\rm 1,3},
    Yusong Deng\textsuperscript{\rm 1,3}
    Meilan Hao\textsuperscript{\rm 1}
}

\affiliations{
    \textsuperscript{\rm 1}AnnLab, Institute of Semiconductor, Chinese Academy of Sciences,
    Beijing, China\\
    \textsuperscript{\rm 2}School of Integrated Circuits, University of Chinese Academy of Sciences,
    Beijing China\\
    \textsuperscript{\rm 3} School of Electronic, Electrical, and Communication Engineering,
University of Chinese Academy of Sciences, 
            Beijing, China\\
    \textsuperscript{\rm 4} Zhongguancun Academy, 
            Beijing, China\\
            
    wjli@semi.ac.cn, yulina@semi.ac.cn, wumin@semi.ac.cn \\
     
}

\usepackage{bibentry}

\begin{document}

\maketitle

\begin{abstract}
Mathematical formulas are the language of communication between humans and nature.
Discovering latent formulas from observed data is an important challenge in artificial intelligence, commonly known as symbolic regression(SR). 
The current mainstream SR algorithms regard SR as a combinatorial optimization problem and use Genetic Programming (GP) or Reinforcement Learning (RL) to solve the SR problem. These methods perform well on simple problems, but poorly on slightly more complex tasks. In addition, this class of algorithms ignores an important aspect: in SR tasks, symbols have explicit numerical meaning. So can we take full advantage of this important property and try to solve the SR problem with more efficient numerical optimization methods? 
The Equation Learner (EQL) replaces activation functions in neural networks with basic symbols and sparsifies connections to derive a simplified expression from a large network. However, EQL's fixed network structure can not adapt to the complexity of different tasks, often resulting in redundancy or insufficient, limiting its effectiveness.
Based on the above analysis, we propose MetaSymNet, a tree-like network that employs the PANGU meta-function as its activation function. PANGU meta-function can evolve into various candidate functions during training. The network structure can also be adaptively adjusted according to different tasks. Then the symbol network evolves into a concise, interpretable mathematical expression. 
To evaluate the performance of MetaSymNet and five baseline algorithms, we conducted experiments across more than ten datasets, including SRBench. 
The experimental results show that MetaSymNet has achieved relatively excellent results on various evaluation metrics.
\end{abstract}

%

\section{Introduction}

Determining a comprehensible and succinct mathematical expression from datasets remains a crucial challenge in artificial intelligence research. Symbolic regression endeavors to identify an interpretable equation $Y=F(X)$ that precisely represents the relationship between the independent variable $X$ and the dependent variable $Y$.
Contemporary predominant approaches to symbolic regression often treat the problem as a combinatorial optimization challenge, wherein each mathematical operator is considered merely an 'action' with no inherent mathematical significance. These methodologies generally employ Genetic Programming (GP) or reinforcement learning techniques. While effective for simpler tasks, their performance deteriorates when addressing complex symbolic regression challenges due to the exponential growth of the search space. On the other hand, in contrast to traditional combinatorial optimization tasks, symbols used in symbolic regression possess inherent mathematical meanings. Consider the classic Traveling Salesman Problem (TSP), where the elements—four cities labeled [A, B, C, D]—lack additional implications beyond their identification. Conversely, in a symbolic regression scenario involving four symbols [+,$\times$, sin,x], these are not merely labels. Each symbol carries a distinct mathematical significance. For instance, these symbols can function as activation functions within neural networks, facilitating processes such as forward or backpropagation.

The Equation Learner (EQL) modifies the architecture of the Multi-Layer Perceptron (MLP) by consistently incorporating a predetermined symbolic function as the activation function in each layer. Following this, a sparsification technique is applied to eliminate redundant connections within the network. This methodological framework allows EQL to derive a concise mathematical expression from complex data relationships. However, EQL has the following problems: \textbf{1) Sparsification Challenges:} Achieving effective sparsification within the fully connected network proves to be more problematic than anticipated. It is often difficult to reduce the network to only one or two connections while maintaining high fitting accuracy solely through $L_1$ regularization. As a result, many times the resulting expression does not fit the data well.
\textbf{2) Fixed Network Structure:} The network structure is fixed and cannot be adjusted according to the complexity of the tasks. It is easy to cause a mismatch between network structure and task. For instance, an excessively large initialized network may produce overly complex expressions, while a network that is too small may compromise fitting performance.

In this paper, we introduce MetaSymNet, an innovative tree-based symbolic regression network. Its architecture is structured as an expression binary tree, featuring compact connections between nodes, which effectively eliminates the need for the sparsification of connection weights. Furthermore, we propose PANGU metafunctions and Variable metafunctions to serve as activation functions for internal nodes and leaf nodes, respectively. The PANGU function possesses the flexibility to evolve into any operational symbol during the training process, while Variable metafunctions enable the selection of the type of input variable. Notably, MetaSymNet is capable of adaptively modifying its structure, allowing it to grow or shrink in response to the complexity of the task at hand.
Our contributions are summarized as follows:
\begin{itemize}

\item We introduce MetaSymNet, an innovative symbolic regression algorithm based on numerical optimization that features dynamic adaptation of node activation functions in response to specific task requirements. Additionally, MetaSymNet's network architecture is capable of real-time adaptive adjustments driven by gradient information, optimizing its structure to align closely with the complexities of the tasks at hand. The source code for MetaSymNet is available \footnote{Code: \url{https://github.com/1716757342/MetSymNet}}.
\item We propose a novel activation function, PANGU meta-function, which can adaptively evolve into various candidate functions during the numerical optimization process.
\item We propose a structural adjustment mechanism that utilizes the difference in the number of inputs required by unary and binary functions. This approach allows MetaSymNet to modify its tree network structure in real time, guided by gradient, progressively refining its topology toward an optimal configuration.

\item We incorporate an entropy loss metric for each set of selected parameters into the loss function. This integration aims to augment MetaSymNet's training efficiency and precision.

\end{itemize}

\section{Related Work}
\textbf{Based on Genetic algorithm}
The genetic algorithm (GA)  \citep{ga,ga2} is a classical algorithm that imitates biological evolution, and the algorithm that applies the GA algorithm to solve the problem of symbolic regression is the Genetic Programming (GP)   \citep{gp1,gp2,gp3} algorithm. GP represents each expression in the form of a binary tree, initializes an expression population, and then evolves a better population employing crossover, mutation, etc. Repeat the process until the termination condition is reached. 
\\
\begin{figure*}[t]
\centering
\vspace{-1.0cm}
\setlength{\belowcaptionskip}{-0.0cm} 
\includegraphics[width=160mm]{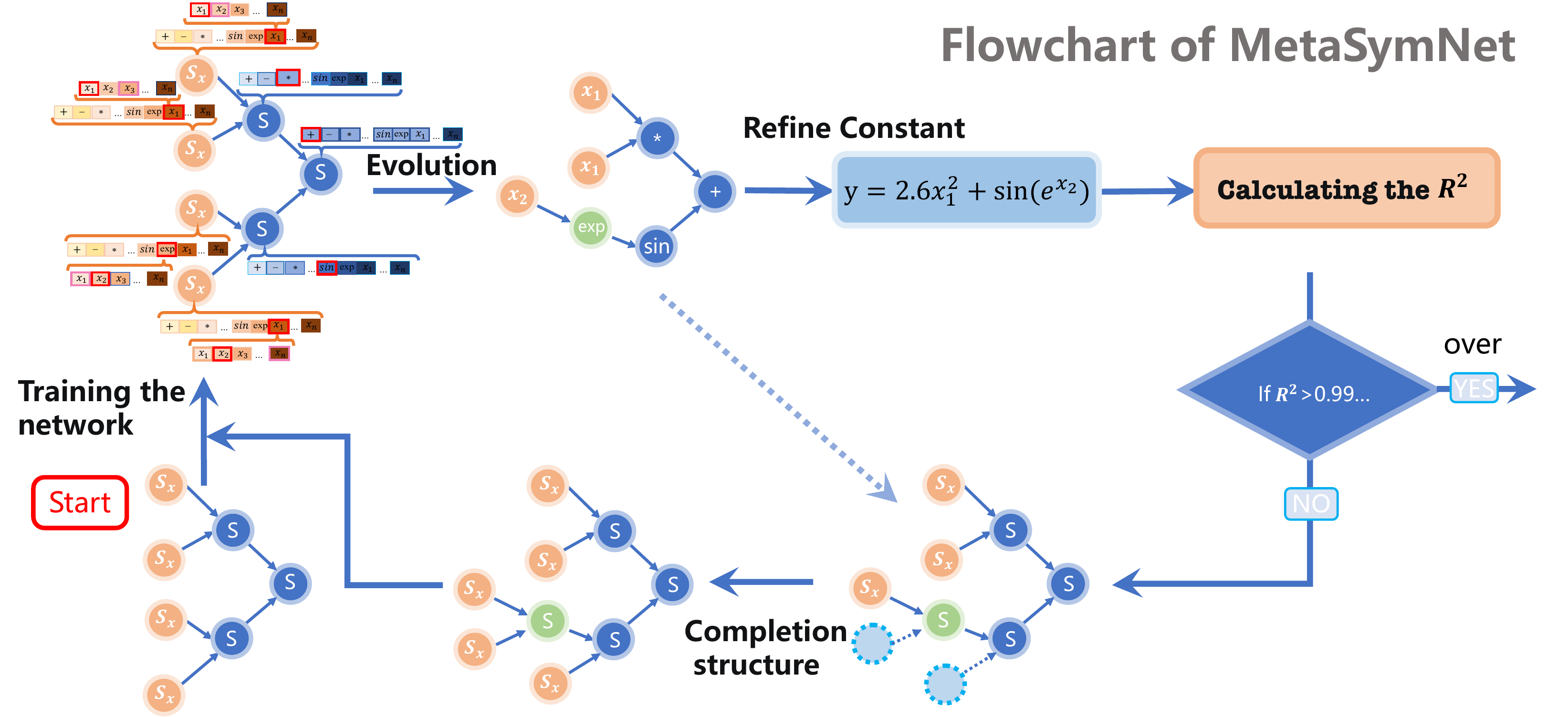}
\caption{Flowchart of the MetaSymNet. (i) First randomly initialize a network, where the internal node $S$ is the PANGU meta-function and the leaf node $S_x$ is the Variable meta-function. (ii) A numerical optimization algorithm is used to optimize the parameters. In this process, the amplitude parameter $\mathcal{W}$ and the bias parameter $\mathcal{B}$ are optimized first, and then the selection parameters $\mathbb{Z}$ and $\mathbb{D}$ of the PANGU meta-functions and Variable meta-functions are optimized. Iterate several times. (iii) After parameter optimization, we determine the basic candidate symbols to which each PANGU metafunction and Variable metafunction should evolve based on the selection parameters $\mathbb{Z}$ and $\mathbb{D}$. (iv) When the network evolves into an expression, we further refine the constants of the expression and calculate the loss and $R^2$. The iteration stops when $ R^2$ reaches the specified threshold. Otherwise, We replace the internal nodes (operation symbols) of the obtained expression binary tree with PANGU metafunctions and leaf nodes (variable symbols) with Variable metafunctions. The network structure is then supplemented with Variable meta-functions such that each PANGU meta-function has two children. The iteration continues. }
\label{fig1}
\end{figure*}
\textbf{Based on reinforcement learning}
Deep Symbolic Regression (DSR) \citep{dsr} is a very good algorithm for symbolic regression using reinforcement learning. DSR uses a recurrent neural network as the strategy network of the algorithm, which takes as input the parent and sibling nodes to be generated, and outputs as the probability of selecting each symbol. DSR uses risk policy gradients to refine policy network parameters. DSO \citep{dso} introduces the GP algorithm based on DSR. This algorithm uses the formula sampled by DSR as the initial population of the GP algorithm and then performs crossover, mutation, and other operations through the GP algorithm to obtain a new population. Then DSO selects the n formulas with the highest reward function and mixes them with the data generated by DSR before, and finally updates the policy network RNN. Then DSR produces a better initial population for GP. Repeat the process.SPL \citep{spl} successfully applies MCTS to solve the problem of symbolic regression. In this algorithm, the author uses MCTS to explore the symbolic space and puts forward a modular concept to improve search efficiency. DySymNet \cite{li2023neural} uses reinforcement learning to guide the generation of symbol networks. RSRM \cite{xu2023rsrm} deeply combines MCTS, Double Q-learning block, and GP, and achieves good performance on many datasets.\\
\textbf{Based on neural networks}
AI Feynman series algorithms are mainly divided into two versions, the main idea of this series of algorithms is to reduce complexity to simplicity. AI Feynman 1.0 \citep{aif1} first uses a neural network to fit the data and then uses the curve fitted by the neural network to analyze a series of properties in the data, such as symmetry and separability. Then the formula to be found is divided into simple units by these properties, and finally, the symbol of each unit is selected by random selection. The idea of AI Feynman 2.0 \citep{aif2} and 1.0 is very similar, the biggest difference between the two is that version 2.0 introduces more properties so that the search expression can be divided into simpler units, improving the search efficiency. Moreover, the application scenario of AI Feynman2.0 is no longer limited to the field of physics, and the application scope is more extensive.
DGP \cite{dgp} works by normalizing the weight connections and then selecting the corresponding symbols. However, it does not solve the problem of variable selection. Very unfriendly to multivariate problems. EQL \citep{eql1,eql2} algorithm is an SR algorithm based on a neural network, which replaces the activation function in the fully connected neural network with basic operation symbols such as $[+, -, ..., sin...]$, then removes the excess connections through pruning, and extracts an expression from the network. \\
\textbf{Based on Transformer}
The NeSymReS \citep{nesy} algorithm treats the symbolic regression problem as a translation problem in which the input $[x,y]$, and the output is a preorder traversal of the expressions. NeSymReS first generates several expressions and then uses the sampled data $[x,y]$ from these expressions as inputs and the backbone of the expressions as outputs to train a transformer \citep{tran}, pre-training model. When predicting the data, $[x,y]$ is entered into the transformer, and then combined with the beam search, a pre-order traversal of the formula is generated in turn. The biggest difference between the end-to-end \citep{end2end} algorithm and NeSymReS is that the end-to-end approach can directly predict a constant to a specific value. Instead of predicting a constant placeholder `C'.  \citep{TPSR} and  \citep{dgsr_mcts}, use the pre-trained model as a policy network to guide the search process of MCTS to improve the search efficiency. The SNIP \citep{meidani2023snip} uses contrastive learning to train the data feature extractor. Then, it trains a transformer to generate the expression skeleton in a self-supervised manner.
\section{Methodology}
MetaSymNet, a tree-like neural network, treats each node as a neuron with internal nodes using the PANGU metafunction as their activation function, and leaf nodes employing Variable metafunctions. During training, in addition to optimizing amplitude parameters $\mathcal{W}$ and bias term $b$ like ordinary fully connected neural networks, we also optimize the internal Selecting parameters $\mathbb{Z}$, $\mathbb{D}$ of the PANGU metafunctions and Variable metafunctions. The end-to-end training of MetaSymNet not only results in a high degree of data fit but also facilitates the evolution of the metafunctions into a variety of fundamental symbols, transforming the network into an analyzable and interpretable mathematical formula.
MetaSymNet's algorithm schematic is shown in Figure \ref{fig1}. See Appendix \ref{pse1} for the pseudocode\\
We begin by stochastically initializing a tree-like network whose architecture and neuronal count are arbitrarily defined. The activation functions of the network are the PANGU meta-functions and Variable metafunctions. In the beginning, each neuron has two inputs because our symbol library contains not only unary activation functions like $[sin, cos,exp, sqrt, log]$ but also binary activation functions like $[+,-,\times,\div]$, etc. 

\subsection{\textbf{PANGU meta-function}}
\begin{figure}
  \centering
  \includegraphics[width=0.42\textwidth]{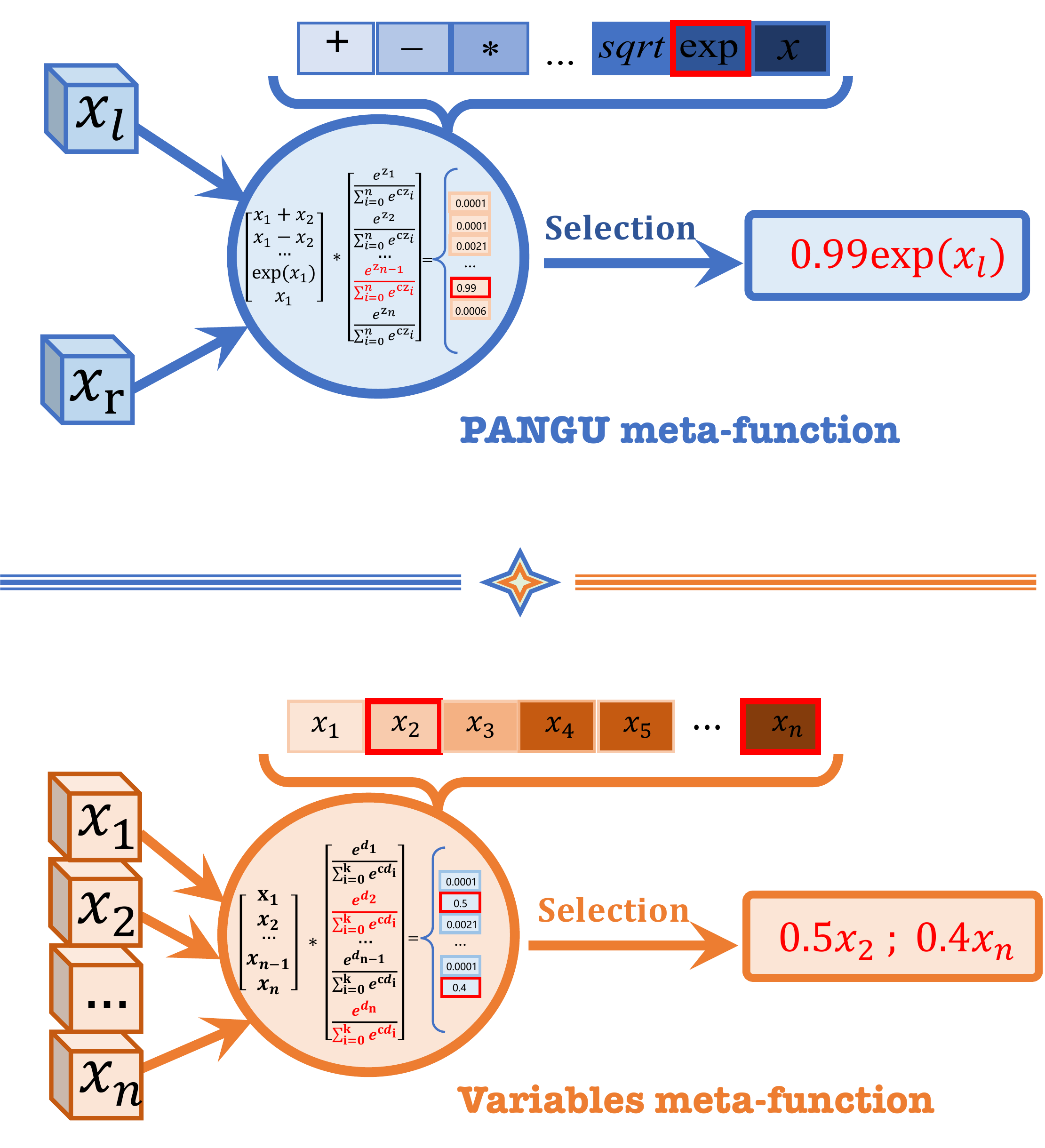}
  \caption{The structure of PANGU metafunctions and Variable metafunctions. The figure depicts the detailed internal structure diagram of the PANGU metafunction (top) and Variable metafunction (bottom). Note: The variable metafunctions are chosen from two variables at a time.}
  \label{fig2}
\end{figure}

In a standard neural network, the activation function is static, typically restricted to a single type such as ReLU or sigmoid. This makes neural networks ultimately a combination and nesting of multiple sets of activation functions and parameters, often a complex and hard-to-interpret `black box'.
Mathematical formulas in natural science are composed of basic operation symbols such as sin() and cos(). Can we design a meta-activation function that can automatically evolve into various basic operators during training? 
To realize this, we design a PANGU meta-function, shown in Fig. \ref{fig2}. The formula is as follows Eq.\ref{e1}:
\begin{align}
\label{e1}
\mathcal{OUT}
&= w * O^T\mathcal{E}
= w * [o_1, o_2, \ldots, o_n]\begin{bmatrix} e_1 \\ e_2 \\ \vdots \\ e_n \end{bmatrix} + b \notag \\
&= w * [x_{l} + x_{r}, \ldots, e^{x_{l}}, x_1, \ldots, x_n]\begin{bmatrix} \frac{e^{z_1}}{\sum_{i=1}^{n} e^{c * z_i}} \\ \frac{e^{z_2}}{\sum_{i=1}^{n} e^{c * z_i}} \\ \ldots \\ \frac{e^{z_n}}{\sum_{i=1}^{n} e^{c * z_i}} \end{bmatrix} + b
\end{align}
Here, $OUT$ is the corresponding output of the PANGU meta-function, $w$, and $b$ are parameters, and $o_i$ is the output of the $i^{th}$ candidate function(e.g., $x_l + x_r$,  sin($x_l$), cos($x_l$),...) in the library.
$\mathcal{E} = softmax(\mathcal{Z})$ (or $\mathcal{E} =softmax(\mathcal{Z}-max(\mathcal{Z}))$ \citep{soft1,soft2}), $e_i = \frac{e^{z_i}}{\sum_{j=1}^{n} e^{c * z_j}}$ is the $i^{th}$ value in vector $\mathcal{E}$. And $e_i$ can as the probability of selecting $i^{th}$ candidate activation function. 
The vector $\mathcal{Z}=[z_1,z_2,...,z_n]$ is a set of selection parameters that can be optimized to control the probability of each activation function being selected. And all internal neuron function selection parameters $\mathcal{Z}$ form the set $\mathbb{Z} = [\mathcal{Z}_1, \mathcal{Z}_2,..., \mathcal{Z}_n]$. Here $n$ denotes the number of internal neurons. 
\subsection{\textbf{Variables meta-function
}}
At the leaf neurons of MetaSymNet, we will initialize a Variable meta-function, shown in Fig \ref{fig2}(bottom), but the function can only evolve into different variables $[x_1,x_2,...,x_n]$. Each time, the top two most likely variables are selected. The specific evolution process is shown in the algorithm \ref{algorithm2}. Its expression is as follows \ref{e3}:\\
\begin{align}
\label{e3}
\mathcal{OUT} &=w * X^T\mathcal{E} + b\notag \\
 &= w * [x_1,x_2,...,x_k]\begin{bmatrix} \frac{e^{d_1}}{\sum_{i=1}^{k} e^{c * d_i}} \\ \frac{e^{d_2}}{\sum_{i=1}^{k} e^{c * d_i}}\\ 
...\\\frac{e^{d_n}}{\sum_{i=1}^{k} e^{c * d_i}}\end{bmatrix} + b
\end{align}
Here, $\mathcal{D} = [d_1,d_2,...,d_k]$ is a set of variable selection parameters, k is the number of variables in the task, and $q$ denotes the total number of leaf nodes. And $w$ is the amplitude control parameter and b is the bias parameter. In MetaSymNet, the variable selection parameter $\mathbb{D} = [\mathcal{D}_1, \mathcal{D}_2,..., \mathcal{D}_q]$ and the activation function selection parameter $\mathbb{Z} $ are optimized together. 
\subsection{\textbf{Parameters optimization
}} 
Each neuron is assigned three types of parameters, an amplitude constant $w$, a bias constant $b$, and a set of selection parameters $\mathcal{Z}=[z_1, z_2,..., z_n]$(The leaf neurons are $\mathcal{D}=[d_1, d_2,..., d_n]$). Here, the amplitude constants $w$ of all neurons form the parameter set $\mathcal{W}$. Similarly, all bias constants $b$ form the set $\mathcal{B}$, and selection parameters $\mathcal{Z}$ for all internal neurons form the set $\mathbb{Z}$. All variable selection parameters form the set $\mathbb{D}$. 
We alternately optimize the above types of parameters with a numerical optimization algorithm (For example, SGD \citep{sgd,sgd2}, BFGS \citep{bfgs}, L-BFGS \citep{lbfgs} etc
). First, the parameters in the set $W$ and $B$ are optimized, and then the parameters in the set $ \mathbb{Z}$ and $ \mathbb{D}$ are optimized, next the activation functions of internal neurons and the variables of the leaf neurons are selected. Finally, after optimizing the parameters alternately for several rounds, we extract a formula from the network and further optimize $W$ and $B$ to achieve higher accuracy.
\subsection{\textbf{Activation function selection
}}
There are many types of candidate activation functions. This article contains four binary activation functions $[+, -, \times, \div]$, five unary activation functions $[sin, cos,exp, log,sqrt]$, and variables $[x_1,x_2,...,x_k]$.

\textbf{PANGU meta-function selection process, }
We initialize a PANGU meta-function with an optimizable selection parameter vector $\mathcal{Z}=[z_1, z_2,..., z_n]$, where $n$ represents the class of candidate functions. We optimize $\mathcal{Z}$ by numerical optimization algorithm, Then the optimized $\mathcal{Z}_ {new} $ each element minus the $Max(Z_ {new}) $, and then sent $softmax$ to get $\mathcal{E} = softmax (\mathcal{Z}_ {new} - Max (\mathcal{Z}_ {new})) = [e_1, e_2,..., e_n] $. Each PANGU meta-function $\mathcal{S}$ has two inputs $ x_l$ and $x_r$ (The output of the two child nodes), and then the candidate functions do the corresponding numerical operations respectively. We can obtain the vector $\mathcal{O}=[x_l + x_r, x_l-x_r, x_r * x_r, x_l/x_r, sin(x_l), cos(x_l), exp(x_l), log(x_l), sqrt(x_l),x_1,\\x_2,...,x_n]$. Finally, we dot multiply the vectors $\mathcal{E}$ and $\mathcal{O}$, then multiply by a constant $w$, and add a bias term $b$ to get the final output. After optimizing the vector $\mathcal{Z}$ multiple times in this way, we map $\mathcal{Z}$ to one-hot \citep{one-hot} form and then dot multiply it with $O$. The final selected activation function is the symbol corresponding to the index equal to 1 in one-hot (Figure \ref{fig1} top left, the evolution of PANGU metafunction $S$). Complete the activation function selection \\
\begin{table*}[t]
    \vspace{-1.1cm}
    \def\arraystretch{1.2}

    \begin{center}
        \bgroup
        \setlength{\tabcolsep}{0.4em}
        \begin{tabular}{c|l|cccccc} 
            \toprule
            \midrule
            \multirow{2}{*}{\bf Group} & \multicolumn{1}{c|}{\multirow{2}{*}{\bf Dataset}} & \multicolumn{6}{c}{\bf BASELINES} \\
            \cline{3-8}
            & & \multicolumn{1}{c}{\bf MetaSymNet} & \multicolumn{1}{c}{\bf DSO} & \multicolumn{1}{c}{\bf TPSR} & \multicolumn{1}{c}{\bf SPL} & \multicolumn{1}{c}{\bf NeSymReS} & \multicolumn{1}{c}{\bf EQL}  \\
            \midrule
            \multirow{10}{*}{\rotatebox{90}{Standards}}             
            & Nguyen    & $\textbf{0.9999}_{\pm0.000}$ & $\textbf{0.9999}_{\pm0.001}$ & $0.9948_{\pm0.003}$& $0.9842_{\pm0.001}$& $0.8468_{\pm0.002}$& $0.9924_{\pm0.005}$  \\
            & Keijzer   & $\textbf{0.9992}_{\pm0.001}$ & $0.9924_{\pm0.001}$& $0.9828_{\pm0.002}$ & $0.8919_{\pm0.002}$ & $0.7992_{\pm0.002}$ & $0.9666_{\pm0.003}$ \\
             & Korns     & $\textbf{0.9999}_{\pm0.001}$  &$ 0.9872_{\pm0.000}$& $0.9325_{\pm0.004}$  & $0.8788_{\pm0.001}$ & $0.8011_{\pm0.001}$  & $0.9285_{\pm0.004}$\\
            & Constant  & $\textbf{0.9996}_{\pm0.002}$  & $0.9988_{\pm0.003}$ & $0.9319_{\pm0.002}$  & $0.8942_{\pm0.002}$  & $0.8444_{\pm0.002}$   & $0.9466_{\pm0.003}$\\
            & Livermore  & $\textbf{0.9924}_{\pm0.003}$ & $0.9746_{\pm0.003}$& $0.8820_{\pm0.004}$& $0.8728_{\pm0.002}$ & $0.7136_{\pm0.004}$  & $0.9037_{\pm0.005}$\\
            & Vladislavleva  & $0.9826_{\pm0.003}$ & $\textbf{0.9963}_{\pm0.004}$ & $0.9128_{\pm0.005}$ & $0.8433_{\pm0.004}$ & $0.6892_{\pm0.004}$  & $0.8926_{\pm0.004}$\\
            & R  & $\textbf{0.9921}_{\pm0.002}$ & $0.9744_{\pm0.003}$ & $0.9422_{\pm0.001}$ & $0.9122_{\pm0.003}$ & $0.8003_{\pm0.004}$  & $0.8637_{\pm0.005}$\\
            & Jin  & $0.9896_{\pm0.002}$ & $\textbf{0.9916}_{\pm0.002}$ & $0.9826_{\pm0.004}$ & $0.9211_{\pm0.002}$& $0.8627_{\pm0.002}$ & $0.9677_{\pm0.004}$\\
             & Neat  & $\textbf{0.9953}_{\pm0.004}$& $ 0.9827_{\pm0.003}$ & $ 0.9319_{\pm0.004}$ & $ 0.8828_{\pm0.003}$& $ 0.7996_{\pm0.005}$ & $0.9631_{\pm0.004}$\\
            & Others  & $\textbf{0.9984}_{\pm0.001}$ & $0.9861_{\pm0.002}$& $0.9667_{\pm0.003}$ & $0.9435_{\pm0.003}$ & $0.8226_{\pm0.002}$  & $0.9438_{\pm0.004}$\\

            \midrule
            \multirow{3}{*}{\rotatebox{90}{SRBench}} 
             & Feynman  & $\textbf{0.9960}_{\pm0.002}$ &$0.9610_{\pm0.003}$ &$0.8928_{\pm0.003}$  & $0.9284_{\pm0.003}$ &$0.7025_{\pm0.003}$ & $0.8725_{\pm0.005}$\\
             & Strogatz  & $\textbf{0.9424}_{\pm0.004}$ &$0.9313_{\pm0.002}$ &$0.8249_{\pm0.003}$  & $0.8411_{\pm0.002}$ &$0.6222_{\pm0.002}$ & $0.8844_{\pm0.005}$\\
             & Black-box & $\textbf{0.9302}_{\pm0.003}$ &$0.9033_{\pm0.004}$ &$0.8753_{\pm0.003}$  & $0.9024_{\pm0.002}$ &$0.6825_{\pm0.003}$ & $0.7852_{\pm0.005}$\\
            \bottomrule
            &Average&\textbf{0.9859} &0.9749  & 0.9024 &0.8997 & 0.7528&0.7852\\
            \bottomrule
        
        \end{tabular}
        \egroup
        \end{center}
        \caption{Comparison of the coefficient of determination ($R^2$) between MetaSymNet and five baseline. Bold values indicate state-of-the-art (SOTA) performance. The confidence level is 0.95.}
        \label{tab1}
\end{table*}
\textbf{Variable meta-function selection process, }
First, suppose that the output of the leaf neuron $S_x$ is $v_i$ after multiple optimizations. Here, $v_i = softmax(D_i) * X$, $X=[x_1,x_2,...,x_n]$. Then the two variables with the highest probability are selected as $x_l$, $x_r$ according to $softmax(D_i)$ (Fig. \ref{fig1} Top left, the first step in the evolution of the Variable metafunction $S_x$, selecting two variables (red and pink boxes) ). Finally, $x_l$ and $x_r$ are used as the input of each candidate activation function, and the $O_i = [x_l + x_r, x_l - x_r,..., sin (x_l),..., x1,..., x_n] $ is obtained, then we choose the symbol corresponding to the value closest to $v_i$ in vector $O_i$ as the new activation function of the leaf neuron (Figure \ref{fig1} top left, the second step in the evolution of Variable metafunction $S_x$). 

If the target expression is still not found, or $R^2$ is less than 0.9999. This process is repeated until a stopping condition is reached.

\subsection{\textbf{Network Structure Optimization
}}
The network structure of traditional neural networks is fixed, which is very easy to cause network structure redundancy. Although there are many pruning methods to simplify the structure of neural networks by pruning unnecessary connections, it is not easy to extract a concise expression from a network  \citep{ac1,ac2}. We propose an algorithm to optimize the MetaSymNet structure in real time under the guidance of gradient. From the above, we know that the activation function of MetaSymNet can be learned dynamically and each neuron has two inputs. We designed a method to realize the dynamic adjustment of the network structure when the activation function changes. The rules for adjusting the structure according to the activation function changes are as follows:\\
\textbf{(1), Structural growth}
\begin{itemize}
    \item When $S_x$ evolves into a unary activation function. At this point, $S_x$ evolves into the chosen unary operator symbol, and, a variable (leaf node) is selected as an input, (For example, the green symbol $exp$ in Fig. \ref{fig1}). Finally, in Figure \ref{fig1} `Completion structure' stage, the unary operation symbol is changed into PANGU meta-function $S$, and two variable meta-functions $S_x$ are added as child nodes. Realize the growth of network structure.
    \item When $S_x$ evolves into a binary activation function(e.g. +, *). At this point, $S_x$ evolves into the chosen binary operator symbol, and two variable leaf nodes are selected as input. Finally, in Figure \ref{fig1} `Completion structure' stage, the binary operation symbol is changed into PANGU meta-function $S$, and two variable meta-functions $S_x$ are added as child nodes. Realize the growth of network structure.
\end{itemize}

\textbf{(2), Structural reduction}
\begin{itemize}
    \item When $S$ evolves into a variable symbol (e.g. $x_1$,$x_n$). At this point, $S$ evolves into the chosen variable symbol, and all child nodes following this node are clipped off. Finally, in Figure \ref{fig1} `Completion structure' stage, the variable symbol is changed into a Variable meta-function $S_x$. Realize the reduction of network structure.
\end{itemize}


\begin{figure*}[!htb] 
    \centering
    \vspace{-1.8cm}
    \begin{subfigure}[b]{0.47\linewidth}
       \includegraphics[width=\linewidth]{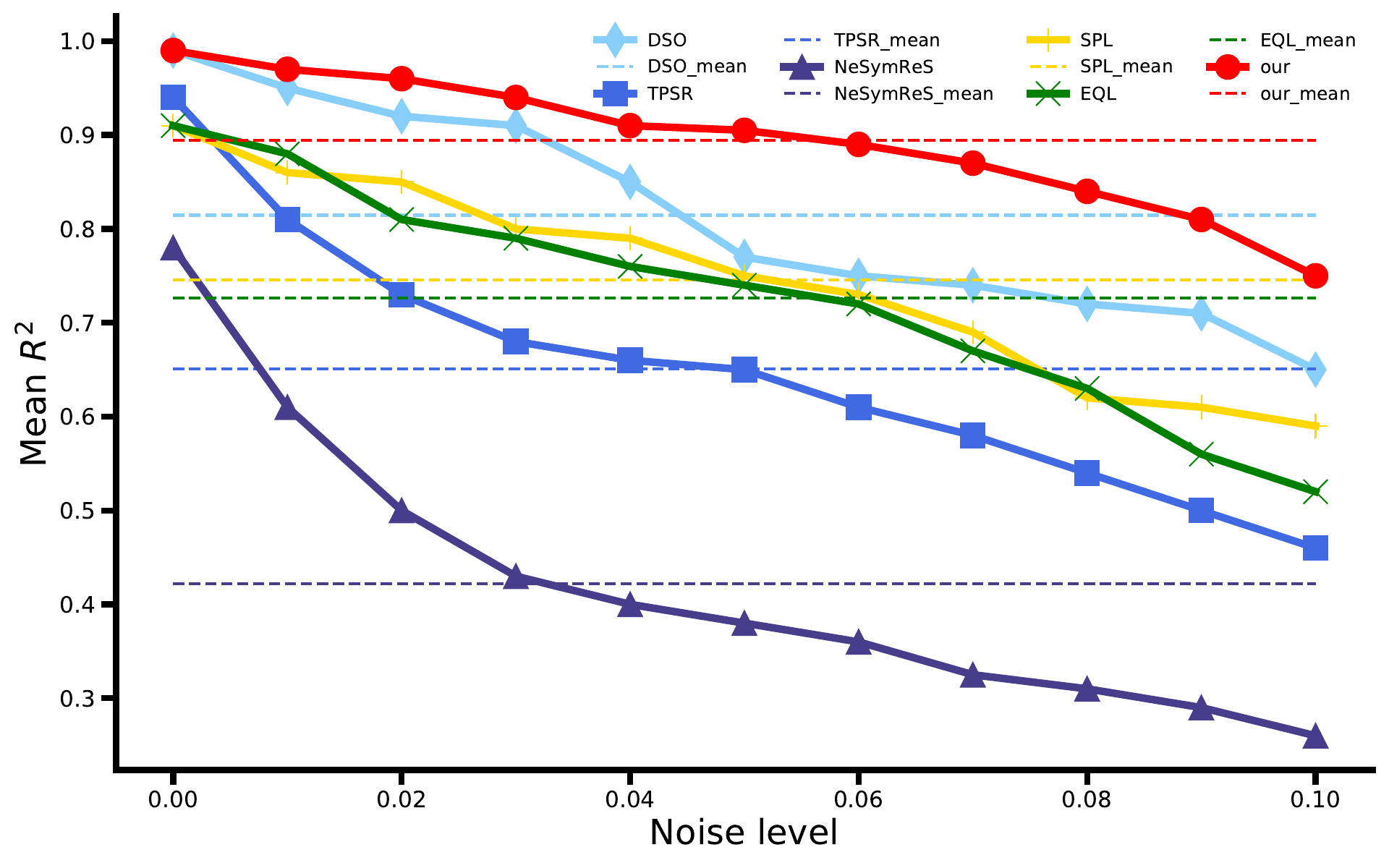}
       \caption{Noise robustness}
       \label{fig3a}
    \end{subfigure}
    \hfill 
    \begin{subfigure}[b]{0.48\linewidth}
        \includegraphics[width=\linewidth]{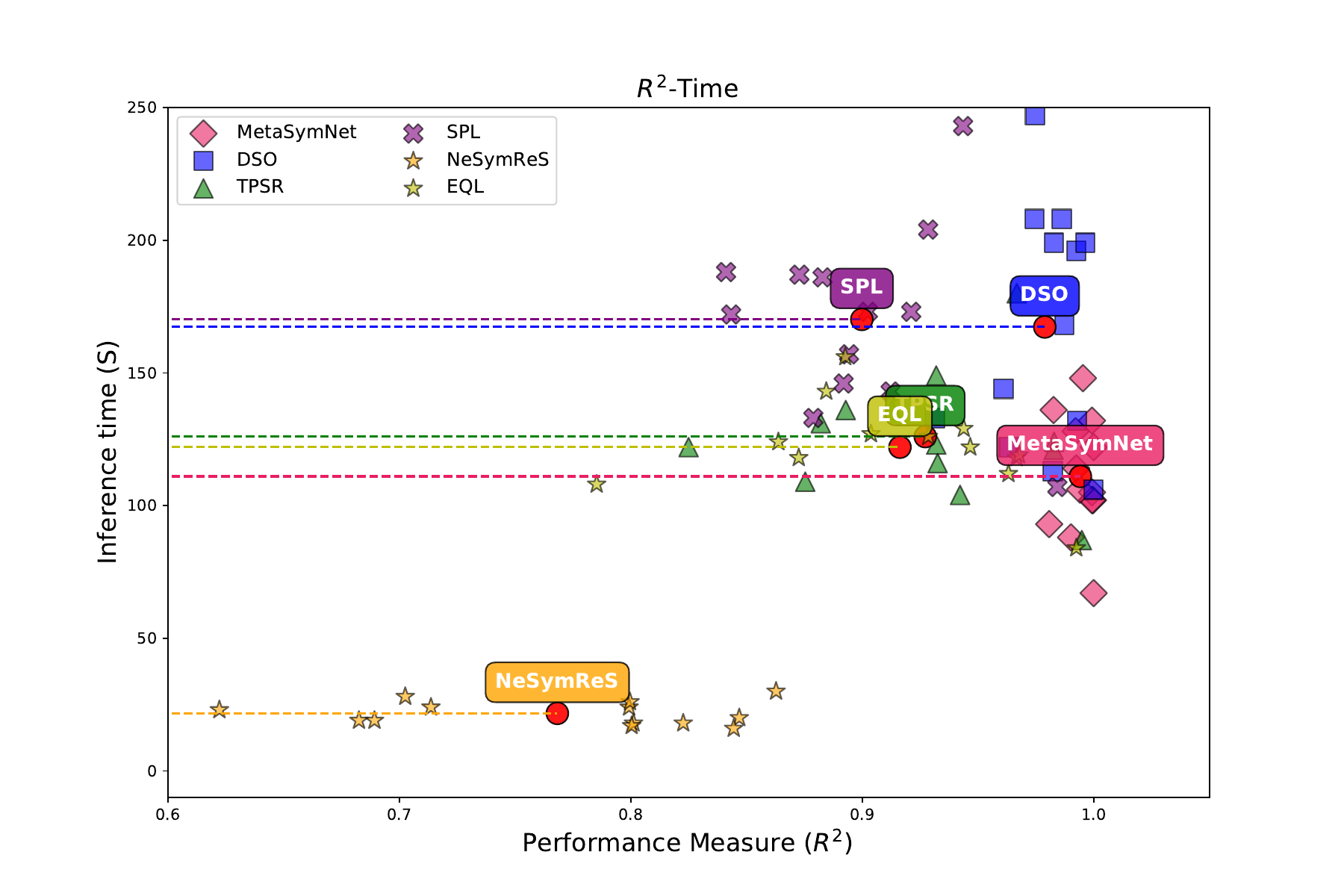}
        \caption{$R^2$-time Pareto plot}
        \label{fig3b}
    \end{subfigure}
    \caption{Analysis on performance. \subref{fig3a} demonstrates the trends of $R^2$ values between MetaSymNet and five baseline methods across varying noise levels. \subref{fig3b} Provides a $R^2$-time Pareto plot of the various algorithms on individual datasets, showing optimization performance.}
    \label{fig3}
\end{figure*}

\subsection{\textbf{Loss function
}}
In the process of MetaSymNet training, the loss function has a crucial position, because it directly determines the direction of neural network optimization. In this study, we introduce a new loss function that aims to further optimize the performance of MetaSymNet. For each PANGU meta-function, it has a selection parameter $\mathcal{Z}$, and $\mathcal{Z}$ gets $\mathcal{E}$ after passing through the softmax function, where $\mathcal{E}=[e_1,e_2,...,e_n]$. We introduce the entropy \citep{entr1,entr2} of $\mathcal{E}$ for all PANGU metafunctions as part of the loss. Our goal is for the largest element of each vector $\mathcal{E}$ to be significantly higher than the others while maintaining a high fitting accuracy. We only have to choose one symbol for the PANGU meta function. To facilitate the evolution of the PANGU meta-function.
Specifically, the expression for the loss function is as follows \ref{e4}:
\begin{align}
\label{e4}
\mathcal{L} & = L_{MSE} + L_{Entr} \notag\\
& = \frac{1}{\mathcal{N}}\sum_{n=1}^{\mathcal{N}}(y_i-\hat{y}_i)^2  - \lambda \frac{1}{\mathcal{M}}\sum_{j=1}^{\mathcal{M}}log(max(\mathcal{E}_j))
\end{align}
Where $\mathcal{N}$ is the number of sample points, y is the true y value, and $\hat{y}$ is the predicted y value. $\mathcal{M}$ is the number of PANGU meta-functions under the current network, and $\mathcal{E}_j$ is the value of the selection parameter of the $j^{th}$ PANGU meta function after passing softmax. $\lambda$ is the entropy loss regulation coefficient.

\section{Results}
In order to test the performance of MetaSymNet, we compare the performance of our algorithm with five state-of-the-art symbolic regression algorithms
(\textbf{DSO \citep{dso}}, \textbf{NeSymReS \citep{nesy}}, \textbf{SPL \citep{spl}}, \textbf{EQL \citep{eql1}}, and \textbf{TPSR \citep{TPSR}}) on ten public datasets $Nguyen$, $Keijzer$, $Korns$, $Constant$, $Livermore$, $R$, $Vladislavlev$, $Jin$, $Neat$, $Others$, and $SRBench$. Specific formula details are given in Table \cref{a-tab1,a-tab2,a-tab3,a-tab5,a-tab6}, in the appendix. Fig. \ref{fig-SRBench}  shows the Pareto plots(time and accuracy) of MetaSymNet and baselines(in SRBench) on black-box data and Feynman. 

\subsection{Fit ability test}
When doing comparative experiments, we strictly control the experimental variables and ensure that all the other conditions are the same except for the algorithm type to ensure the fairness and credibility of the experiment. Specifically, (1) At test time, on the same test expression, we sample $x_i$ on the same interval for different algorithms. To ensure the consistency of the test data. See Appendix \cref{a-tab1,a-tab2,a-tab3,a-tab5,a-tab6},  for the specific sampling range of each expression. (2) For some of the `Constraints' used in our algorithm, we also use the same `Constraints' in other algorithms. To ensure that the performance of the individual algorithms is not affected by the difference in the `Constraints'. In the experiment, we use the coefficient of determination ($R^2$) \citep{r21,r22} to judge how well each algorithm fits the expression. The results are shown in Table \ref{tab1}, from which we see that MetSymNet achieves the performance of SOTA on multiple datasets. The expression for $R^2$ is given in the following: $\mathcal{R}^2 = 1-\frac{\sum_{i=1}^{\mathcal{N}}(y-\hat{y})^2}{\sum_{i=1}^{\mathcal{N}}(y- \overline{y})^2}$; 
Where $\hat{y}$ is the predicted y value and $\overline{y}$ is the mean of the true y.
\subsection{Noise robustness test } 
In the real world, data often contains noise and uncertainty. This noise can come from measurement errors, interference during data acquisition, or other unavoidable sources \citep{no1,no2,no3}.
The Noise robustness test can simulate the noise situation in the real environment, and help to evaluate the performance of the symbolic regression algorithm in the case of imperfect data, to help us understand its reliability and robustness in practical applications. These trials can guide the tuning and selection of the model, ensuring that it will work robustly in real scenarios. \citep{ano1,ano2}. We generate noisy data $y_{noise}$ in the following way. To simulate different levels of noise in the real world, we divide the noise into ten levels.
First, the noisy data $Data_{noise}$ is obtained by randomly sampling on the interval $[-\mathcal{L} *Span, +\mathcal{L}*Span]$. Here, $\mathcal{L}=[0.00.0.01.0.02,...,0.1]$ is the level of noise. $Span = abs | max(y) - min (y)|$ is the Span of $y$. $y_{noise} = y + Data_{noise}$. We employed the datasets to assess the algorithm's noise robustness. For each mathematical expression, we conducted 20 times at varying noise levels.
Subsequently, we computed the $R^2$ between the curve derived from each trial and the original curve, serving as a metric to quantify noise resilience. The outcome was determined by averaging the results from the 20 trials.
The average $R^2$ of both MetaSymNet and five symbolic regression baselines were compared across different noise levels, and the results are depicted in Figure \ref{fig3a}. 
The comprehensive analysis demonstrates that MetaSymNet exhibits superior noise immunity performance in contrast to the other five symbolic regression baselines.
\subsection{Inference time test}
In order to evaluate a symbolic regression algorithm, in addition to fitting ability and anti-noise ability, the inference speed of the algorithm is also an extremely important index. Therefore, in order to test the inference efficiency of MetaSymNet and the various baselines, we picked the part of the expression data in all test datasets where the various algorithms could achieve $R^2>0.99$. These selected expressions make up dataset $\mathbb{A}$. We then test each expression in dataset $\mathbb{A}$ 10 times with each algorithm. We plot the $R^2-time$ coordinate plot as shown in Fig \ref{fig3},  For this plot, each color corresponds to an algorithm, and each scatter corresponds to the average $R^2$ and average inference time for all expressions in one test dataset. The red dot indicates the center of gravity of each algorithm, and the closer it is to the bottom-right corner of the figure, the better the overall performance of the algorithm. The red star part represents our algorithm. From the figure, we can see that MetaSymNet achieves a good balance between efficiency and accuracy. In addition, Fig\ref{fig-SRBench} shows the graph of MetaSymNet compared with 20 other baselines under the SRBench standard. We believe that the reason for MetaSymNet's efficiency is that the symbolic regression is changed from a combinatorial optimization problem to a numerical optimization problem while retaining the binary tree representation of the expression. As we know, in the current field of machine learning, the efficiency of numerical optimization algorithms is higher than that of reinforcement learning and evolutionary computation algorithms. Therefore, compared with the traditional symbolic regression algorithm, MetaSymNet is more efficient.
\begin{table}
\center
\vspace{-1.2cm}
\resizebox{8.5cm}{!}{
\begin{tabular}{ccccccc}
\toprule[1.45pt]
 Dataset       & MetaSymNet & DSO   & SPL   & TPSR & NeSymReS & EQL  \\ 
 \toprule
 Nguyen        & 16.5       & 20.3  & 26.0  & $\textbf{16.0}$  & 18.2  & 22.4\\
 Keijzer       & $\textbf{18.0}$       & 18.4  & 28.4  & 20.6  & 21.3  & 20.8\\   
 Constant      & 24.4       & 26.6  & 33.5  & $\textbf{22.9}$  & 24.1  & 32.9\\
 Livermore     & 34.8       & 38.2  & 47.3  & 35.3  & $\textbf{32.9}$  & 41.5\\   
 Vladislavleva & 42.4       & 46.3  & 59.4  & 38.2  & $\textbf{36.2}$  & 49.2\\
 R             & 28.5       & 31.3  & 38.2  & $\textbf{24.6}$  & 27.3  & 36.9\\
 Jin           & 20.6       & 22.0  & 32.0  & $\textbf{16.2}$  & 19.9  & 28.4\\
 Others        & $\textbf{28.3}$       & 33.2  & 39.5  & 29.5  & 32.2  & 37.4\\    
 Neat          & 19.5       & 22.7  & 28.7  & $\textbf{16.4}$  & 20.6  & 27.2\\
 Korns         & 25.8       & 26.8  & 32.4  & $\textbf{22.5}$  & 23.5  & 32.4\\
 Feynman       & 23.1       & 24.1  & 34.2  & $\textbf{21.3}$  & 22.4  & 26.6\\   
 Strogatz      & $\textbf{21.4}$       & 27.2  & 32.3  & 24.4  & 28.1  & 31.9\\  
 Black-box     & $\textbf{25.5}$       & 32.9  & 39.2  & 29.3  & 33.9  & 35.3\\  
\toprule
 Average       & 25.3       & 28.5  & 33.2  & $\textbf{24.4}$  & 26.2  & 32.5\\
\toprule
\end{tabular}
}
\caption{Comparison of the average number of symbols (complexity) of the resulting expressions between MetaSymNet and the other five baselines.
}
\label{tab: node}
\end{table}

\subsection{Result complexity test}
In symbolic regression, our ultimate goal is to get a relatively concise expression to fit the observed data. If the resulting expression is too complex, then its interpretability is greatly reduced. Therefore, we set up the following experiment to compare the complexity  (the number of nodes in the binary tree) of the resulting expressions obtained by each algorithm. we selected the expression in which all methods in the database can achieve $R^2>0.999$ as the test set to compare the complexity (number of nodes) of the expression obtained by different algorithms when  $R^2>0.999$. Each expression was run 20 times and then averaged. The maximum length is set to 80 for all algorithms.
The specific statistical results are shown in Table \ref{tab: node}. As we can see from the table, MetaSymNet has the second-lowest average number of nodes after TPSR.
\subsection{Ablation experiments with entropy loss}
For each PANGU metafunction, we have a set of selection $\mathcal{Z}$ passes through $softmax()$ to obtain $\mathcal{E}=[e_1,e_2,...,e_n]$, here $\sum_{i=1}^{i=n}e_i=1$.
To improve the efficiency and performance of MetaSymNet, we introduce the entropy of $\mathcal{E}$ of all PANGU meta-functions in the network as part of the loss function. Specifically $L_{Entr}=\frac{1}{\mathcal{M}}\sum_{j=1}^{\mathcal{M}}log(max(\mathcal{E}_j))$, here, $\mathcal{M}$ denotes the number of PANGU metafunctions in the network. To demonstrate the effectiveness of entropy loss, we perform ablation experiments on it. The specific experimental results are shown in Table \ref{tab-entr}. We think that the introduction of entropy loss can promote the values in $\mathcal{E}$ to appear as `big is bigger, small is smaller'. Makes One of its values significantly larger than the others, which is closer to the one-hot form. It promotes a more efficient and accurate evolution of the PANGU meta-function to different activation functions.

\begin{table}[!t]
\center
\vspace{-1.2cm}
\resizebox{90mm}{!}{
\begin{tabular}{ccccccc}
\toprule[1.45pt]
Data& \multicolumn{3}{c}{MetaSymNet(With entropy loss)}& \multicolumn{3}{c}{MetaSymNet(Without entropy loss)}\\  
\cmidrule[0.1pt]{1-7}
&$R^2 \uparrow$  &Nodes $\downarrow$ & Time(s)$\downarrow$  & $R^2 \uparrow$ &Nodes $\downarrow$ & Time(s) $\downarrow$ \\ 
\cmidrule(lr){1-1}
\cmidrule(lr){2-4}
\cmidrule(lr){5-7}
Nguyen & $0.9999_{\pm0.001} $   &16.5 &96 &$0.9998_{\pm0.04}$  &18.7 &118 \\
Keijzer & $ 0.9992_{\pm0.001}$  &18.0&124  & $0.9977_{\pm0.03} $&23.8 &135\\
Korns &  $ 0.9999_{\pm0.001}$   &24.4 &103&$ 0.9924_{\pm0.06} $&27.3 &132\\
Constant & $  0.9996_{\pm0.002} $ &34.8 &109 &$ 0.9872_{\pm0.04} $&38.3 &124\\
Livermore & $  0.9924_{\pm0.003} $ &42.4 &104 &$0.9834_{\pm0.05}$  &49.2 &122 \\
Vladislavleva & $ 0.9826_{\pm0.003}$ &28.5 &122&$ 0.9807_{\pm0.06}$ &32.2 &132  \\
R & $  0.9921_{\pm0.002} $ &20.6 &100 &$ 0.9829_{\pm0.05}$  &26.4 &111 \\
Jin & $  0.9896_{\pm0.002}$  &28.3 &112 &$ 0.9744_{\pm0.06}$  &32.9 &125\\
Neat & $  0.9953_{\pm0.004}$   &19.5 &104 & $0.9881_{\pm0.05}$ &24.7 &131\\
Others &  $0.9984_{\pm0.001}$  &25.8 &121& $0.9904_{\pm0.05}$   &31.2 &136\\
Feynman &  $0.9960_{\pm0.002} $& 23.1 & 128 & $0.9763_{\pm0.04}$ &27.4 &142\\
Strogatz& $0.9424_{\pm0.004}$ &21.4 &132 &$ 0.9114_{\pm0.06}$ &28.2 &151\\ 
Black-box& $0.9302_{\pm0.003}$ &25.5 &142 &$ 0.9006_{\pm0.07}$ &26.3 &164\\
\cmidrule(lr){1-1}
\cmidrule(lr){2-4}
\cmidrule(lr){5-7}
Average & 0.9859 &25.3 &115& 0.9743 &29.7 &133\\
\bottomrule
\end{tabular}
}
\caption{Ablation experiments on whether to introduce entropy loss in the loss function for MetaSymNet.
\label{tab-entr}}
\end{table}
\section{Discussion and Conclusion}
In this paper, we propose MetaSymNet, which treats the SR as a numerical optimization problem rather than a combinatorial optimization problem. MetaSymNet's structure is a tree-like network that is dynamically adjusted during training and can be expanded or reduced. Compared with the baselines, MetaSymNet has a better fitting ability, noise robustness, and complexity. We propose a PANGU meta-function as the activation function of MetaSymNet. The function can autonomously evolve into various candidate functions under the control of selection parameters. In addition, we present variable metafunctions that can be used to select variables.
Furthermore, the final result of MetaSymNet is a concise, interpretable expression. This characteristic enhances the credibility of MetaSymNet and presents significant potential for application in fields that involve high-risk decision-making, such as finance, medicine, and law. In such domains, where decisions can profoundly impact people's lives, people must understand and trust the algorithm's decision-making process. 
Despite MetaSymNet yielding satisfactory results, it has its limitations. For instance, tuning certain hyperparameters, such as the $\lambda$ in the loss function, proves to be challenging. Additionally, the method can occasionally become trapped in local optima, resulting in approximate rather than exact expressions. Next, we plan to alter the evolution process of PANGU metafunctions. Specifically, instead of relying on the greedy strategy for function selection, we intend to explore a variety of search methods, including beam search, Monte Carlo Tree Search, and others, to enhance the algorithm's performance.
\section {Acknowledgements}
This work was supported in part by the National Natural Science Foundation of China under Grant 92370117, in part by CAS Project for Young Scientists in Basic Research under Grant YSBR-090

\bigskip

\bibliography{aaai25}

\begin{thebibliography}{43}
\providecommand{\natexlab}[1]{#1}

\bibitem[{Augusto and Barbosa(2000)}]{gp3}
Augusto, D.~A.; and Barbosa, H.~J. 2000.
\newblock Symbolic regression via genetic programming.
\newblock In \emph{Proceedings. Vol. 1. Sixth Brazilian symposium on neural networks}, 173--178. IEEE.

\bibitem[{Beall and Lowe(2007)}]{no3}
Beall, E.~B.; and Lowe, M.~J. 2007.
\newblock Isolating physiologic noise sources with independently determined spatial measures.
\newblock \emph{Neuroimage}, 37(4): 1286--1300.

\bibitem[{Berglund, Hassmen, and Job(1996)}]{no1}
Berglund, B.; Hassmen, P.; and Job, R.~S. 1996.
\newblock Sources and effects of low-frequency noise.
\newblock \emph{The Journal of the Acoustical Society of America}, 99(5): 2985--3002.

\bibitem[{Biggio et~al.(2021)Biggio, Bendinelli, Neitz, Lucchi, and Parascandolo}]{nesy}
Biggio, L.; Bendinelli, T.; Neitz, A.; Lucchi, A.; and Parascandolo, G. 2021.
\newblock Neural symbolic regression that scales.
\newblock In \emph{International Conference on Machine Learning}, 936--945. PMLR.

\bibitem[{Bottou(2010)}]{sgd2}
Bottou, L. 2010.
\newblock Large-scale machine learning with stochastic gradient descent.
\newblock In \emph{Proceedings of COMPSTAT'2010: 19th International Conference on Computational StatisticsParis France, August 22-27, 2010 Keynote, Invited and Contributed Papers}, 177--186. Springer.

\bibitem[{Bottou(2012)}]{sgd}
Bottou, L. 2012.
\newblock Stochastic gradient descent tricks.
\newblock In \emph{Neural Networks: Tricks of the Trade: Second Edition}, 421--436. Springer.

\bibitem[{Castellano, Fanelli, and Pelillo(1997)}]{prun}
Castellano, G.; Fanelli, A.~M.; and Pelillo, M. 1997.
\newblock An iterative pruning algorithm for feedforward neural networks.
\newblock \emph{IEEE transactions on Neural networks}, 8(3): 519--531.

\bibitem[{Dai(2002)}]{bfgs}
Dai, Y.-H. 2002.
\newblock Convergence properties of the BFGS algoritm.
\newblock \emph{SIAM Journal on Optimization}, 13(3): 693--701.

\bibitem[{Dong, Zhu, and Ma(2019)}]{soft2}
Dong, X.; Zhu, X.; and Ma, D. 2019.
\newblock Hardware implementation of softmax function based on piecewise LUT.
\newblock In \emph{2019 IEEE International Workshop on Future Computing (IWOFC}, 1--3. IEEE.

\bibitem[{Drucker et~al.(1996)Drucker, Burges, Kaufman, Smola, and Vapnik}]{svr}
Drucker, H.; Burges, C.~J.; Kaufman, L.; Smola, A.; and Vapnik, V. 1996.
\newblock Support vector regression machines.
\newblock \emph{Advances in neural information processing systems}, 9.

\bibitem[{Espejo, Ventura, and Herrera(2009)}]{gp1}
Espejo, P.~G.; Ventura, S.; and Herrera, F. 2009.
\newblock A survey on the application of genetic programming to classification.
\newblock \emph{IEEE Transactions on Systems, Man, and Cybernetics, Part C (Applications and Reviews)}, 40(2): 121--144.

\bibitem[{Fortin et~al.(2012)Fortin, De~Rainville, Gardner, Parizeau, and Gagn{\'e}}]{gp2}
Fortin, F.-A.; De~Rainville, F.-M.; Gardner, M.-A.~G.; Parizeau, M.; and Gagn{\'e}, C. 2012.
\newblock DEAP: Evolutionary algorithms made easy.
\newblock \emph{The Journal of Machine Learning Research}, 13(1): 2171--2175.

\bibitem[{Gao et~al.(2020)Gao, Saha, Prasad, and Roychoudhury}]{ano2}
Gao, X.; Saha, R.~K.; Prasad, M.~R.; and Roychoudhury, A. 2020.
\newblock Fuzz testing based data augmentation to improve robustness of deep neural networks.
\newblock In \emph{Proceedings of the acm/ieee 42nd international conference on software engineering}, 1147--1158.

\bibitem[{Hermundstad et~al.(2011)Hermundstad, Brown, Bassett, and Carlson}]{ac1}
Hermundstad, A.~M.; Brown, K.~S.; Bassett, D.~S.; and Carlson, J.~M. 2011.
\newblock Learning, memory, and the role of neural network architecture.
\newblock \emph{PLoS computational biology}, 7(6): e1002063.

\bibitem[{Kamienny et~al.(2022)Kamienny, d'Ascoli, Lample, and Charton}]{end2end}
Kamienny, P.-A.; d'Ascoli, S.; Lample, G.; and Charton, F. 2022.
\newblock End-to-end symbolic regression with transformers.
\newblock \emph{Advances in Neural Information Processing Systems}, 35: 10269--10281.

\bibitem[{Kamienny et~al.(2023)Kamienny, Lample, Lamprier, and Virgolin}]{dgsr_mcts}
Kamienny, P.-A.; Lample, G.; Lamprier, S.; and Virgolin, M. 2023.
\newblock Deep Generative Symbolic Regression with Monte-Carlo-Tree-Search.
\newblock \emph{arXiv preprint arXiv:2302.11223}.

\bibitem[{Katoch, Chauhan, and Kumar(2021)}]{ga2}
Katoch, S.; Chauhan, S.~S.; and Kumar, V. 2021.
\newblock A review on genetic algorithm: past, present, and future.
\newblock \emph{Multimedia tools and applications}, 80: 8091--8126.

\bibitem[{Kim et~al.(2020)Kim, Lu, Mukherjee, Gilbert, Jing, {\v{C}}eperi{\'c}, and Solja{\v{c}}i{\'c}}]{eql2}
Kim, S.; Lu, P.~Y.; Mukherjee, S.; Gilbert, M.; Jing, L.; {\v{C}}eperi{\'c}, V.; and Solja{\v{c}}i{\'c}, M. 2020.
\newblock Integration of neural network-based symbolic regression in deep learning for scientific discovery.
\newblock \emph{IEEE transactions on neural networks and learning systems}, 32(9): 4166--4177.

\bibitem[{Li et~al.(2023)Li, Li, Yu, Wu, Liu, and Li}]{li2023neural}
Li, W.; Li, W.; Yu, L.; Wu, M.; Liu, J.; and Li, Y. 2023.
\newblock A Neural-Guided Dynamic Symbolic Network for Exploring Mathematical Expressions from Data.
\newblock \emph{arXiv preprint arXiv:2309.13705}.

\bibitem[{Liu and Nocedal(1989)}]{lbfgs}
Liu, D.~C.; and Nocedal, J. 1989.
\newblock On the limited memory BFGS method for large scale optimization.
\newblock \emph{Mathematical programming}, 45(1-3): 503--528.

\bibitem[{Maksimenko et~al.(2018)Maksimenko, Kurkin, Pitsik, Musatov, Runnova, Efremova, Hramov, and Pisarchik}]{ac2}
Maksimenko, V.~A.; Kurkin, S.~A.; Pitsik, E.~N.; Musatov, V.~Y.; Runnova, A.~E.; Efremova, T.~Y.; Hramov, A.~E.; and Pisarchik, A.~N. 2018.
\newblock Artificial neural network classification of motor-related eeg: An increase in classification accuracy by reducing signal complexity.
\newblock \emph{Complexity}, 2018.

\bibitem[{Martius and Lampert(2016)}]{eql1}
Martius, G.; and Lampert, C.~H. 2016.
\newblock Extrapolation and learning equations.
\newblock \emph{arXiv preprint arXiv:1610.02995}.

\bibitem[{Matsubara et~al.(2022)Matsubara, Chiba, Igarashi, and Ushiku}]{matsubara2022rethinking}
Matsubara, Y.; Chiba, N.; Igarashi, R.; and Ushiku, Y. 2022.
\newblock Rethinking symbolic regression datasets and benchmarks for scientific discovery.
\newblock \emph{arXiv preprint arXiv:2206.10540}.

\bibitem[{Meidani et~al.(2023)Meidani, Shojaee, Reddy, and Farimani}]{meidani2023snip}
Meidani, K.; Shojaee, P.; Reddy, C.~K.; and Farimani, A.~B. 2023.
\newblock SNIP: Bridging Mathematical Symbolic and Numeric Realms with Unified Pre-training.
\newblock \emph{arXiv preprint arXiv:2310.02227}.

\bibitem[{Mirjalili and Mirjalili(2019)}]{ga}
Mirjalili, S.; and Mirjalili, S. 2019.
\newblock Genetic algorithm.
\newblock \emph{Evolutionary Algorithms and Neural Networks: Theory and Applications}, 43--55.

\bibitem[{Mundhenk et~al.(2021)Mundhenk, Landajuela, Glatt, Santiago, Faissol, and Petersen}]{dso}
Mundhenk, T.~N.; Landajuela, M.; Glatt, R.; Santiago, C.~P.; Faissol, D.~M.; and Petersen, B.~K. 2021.
\newblock Symbolic regression via neural-guided genetic programming population seeding.
\newblock \emph{arXiv preprint arXiv:2111.00053}.

\bibitem[{Nagelkerke et~al.(1991)}]{r21}
Nagelkerke, N.~J.; et~al. 1991.
\newblock A note on a general definition of the coefficient of determination.
\newblock \emph{biometrika}, 78(3): 691--692.

\bibitem[{Ozer(1985)}]{r22}
Ozer, D.~J. 1985.
\newblock Correlation and the coefficient of determination.
\newblock \emph{Psychological bulletin}, 97(2): 307.

\bibitem[{Petersen et~al.(2019)Petersen, Landajuela, Mundhenk, Santiago, Kim, and Kim}]{dsr}
Petersen, B.~K.; Landajuela, M.; Mundhenk, T.~N.; Santiago, C.~P.; Kim, S.~K.; and Kim, J.~T. 2019.
\newblock Deep symbolic regression: Recovering mathematical expressions from data via risk-seeking policy gradients.
\newblock \emph{arXiv preprint arXiv:1912.04871}.

\bibitem[{R{\'e}nyi(1961)}]{entr2}
R{\'e}nyi, A. 1961.
\newblock On measures of entropy and information.
\newblock In \emph{Proceedings of the Fourth Berkeley Symposium on Mathematical Statistics and Probability, Volume 1: Contributions to the Theory of Statistics}, volume~4, 547--562. University of California Press.

\bibitem[{Rodr{\'\i}guez et~al.(2018)Rodr{\'\i}guez, Bautista, Gonzalez, and Escalera}]{one-hot}
Rodr{\'\i}guez, P.; Bautista, M.~A.; Gonzalez, J.; and Escalera, S. 2018.
\newblock Beyond one-hot encoding: Lower dimensional target embedding.
\newblock \emph{Image and Vision Computing}, 75: 21--31.

\bibitem[{Rumelhart, Hinton, and Williams(1986)}]{bpnn}
Rumelhart, D.~E.; Hinton, G.~E.; and Williams, R.~J. 1986.
\newblock Learning representations by back-propagating errors.
\newblock \emph{nature}, 323(6088): 533--536.

\bibitem[{Shojaee et~al.(2023)Shojaee, Meidani, Farimani, and Reddy}]{TPSR}
Shojaee, P.; Meidani, K.; Farimani, A.~B.; and Reddy, C.~K. 2023.
\newblock Transformer-based Planning for Symbolic Regression.
\newblock \emph{arXiv preprint arXiv:2303.06833}.

\bibitem[{Stevens et~al.(2021)Stevens, Venkatesan, Dai, Khailany, and Raghunathan}]{soft1}
Stevens, J.~R.; Venkatesan, R.; Dai, S.; Khailany, B.; and Raghunathan, A. 2021.
\newblock Softermax: Hardware/software co-design of an efficient softmax for transformers.
\newblock In \emph{2021 58th ACM/IEEE Design Automation Conference (DAC)}, 469--474. IEEE.

\bibitem[{Sun et~al.(2022)Sun, Liu, Wang, and Sun}]{spl}
Sun, F.; Liu, Y.; Wang, J.-X.; and Sun, H. 2022.
\newblock Symbolic physics learner: Discovering governing equations via monte carlo tree search.
\newblock \emph{arXiv preprint arXiv:2205.13134}.

\bibitem[{Tam et~al.(2008)Tam, Viswanathan, Ahuja, and Panda}]{no2}
Tam, C.~K.; Viswanathan, K.; Ahuja, K.; and Panda, J. 2008.
\newblock The sources of jet noise: experimental evidence.
\newblock \emph{Journal of Fluid Mechanics}, 615: 253--292.

\bibitem[{Udrescu et~al.(2020)Udrescu, Tan, Feng, Neto, Wu, and Tegmark}]{aif2}
Udrescu, S.-M.; Tan, A.; Feng, J.; Neto, O.; Wu, T.; and Tegmark, M. 2020.
\newblock AI Feynman 2.0: Pareto-optimal symbolic regression exploiting graph modularity.
\newblock \emph{Advances in Neural Information Processing Systems}, 33: 4860--4871.

\bibitem[{Udrescu and Tegmark(2020)}]{aif1}
Udrescu, S.-M.; and Tegmark, M. 2020.
\newblock AI Feynman: A physics-inspired method for symbolic regression.
\newblock \emph{Science Advances}, 6(16): eaay2631.

\bibitem[{Vaswani et~al.(2017)Vaswani, Shazeer, Parmar, Uszkoreit, Jones, Gomez, Kaiser, and Polosukhin}]{tran}
Vaswani, A.; Shazeer, N.; Parmar, N.; Uszkoreit, J.; Jones, L.; Gomez, A.~N.; Kaiser, {\L}.; and Polosukhin, I. 2017.
\newblock Attention is all you need.
\newblock \emph{Advances in neural information processing systems}, 30.

\bibitem[{Wehrl(1978)}]{entr1}
Wehrl, A. 1978.
\newblock General properties of entropy.
\newblock \emph{Reviews of Modern Physics}, 50(2): 221.

\bibitem[{Xu, Liu, and Sun(2023)}]{xu2023rsrm}
Xu, Y.; Liu, Y.; and Sun, H. 2023.
\newblock RSRM: Reinforcement Symbolic Regression Machine.
\newblock \emph{arXiv preprint arXiv:2305.14656}.

\bibitem[{Zeng et~al.(2023)Zeng, Song, Lensen, Ou, Sun, Zhang, and Lv}]{dgp}
Zeng, P.; Song, X.; Lensen, A.; Ou, Y.; Sun, Y.; Zhang, M.; and Lv, J. 2023.
\newblock Differentiable Genetic Programming for High-dimensional Symbolic Regression.
\newblock \emph{arXiv preprint arXiv:2304.08915}.

\bibitem[{Ziyadinov and Tereshonok(2022)}]{ano1}
Ziyadinov, V.; and Tereshonok, M. 2022.
\newblock Noise immunity and robustness study of image recognition using a convolutional neural network.
\newblock \emph{Sensors}, 22(3): 1241.

\end{thebibliography}

\appendix
\newpage
\onecolumn
\appendix

\section*{\huge Appendix for ``MetaSymNet: A Tree-like Symbol Network with Adaptive Architecture and Activation Functions''}

\vspace{10mm} 
\subsection*{\LARGE  List of supplementary materials}
\subsection*{\large Supplementary material on the methodology:}
\begin{itemize}
    \item \ref{AA} Appendix: Pseudocode for the MetaSymNet.
    \item \ref{AB} Appendix: Pseudocode for how to evolve into expressions.
    \item \ref{AC} Appendix: Backpropagation derivation (Use SGD).
    \item \ref{AD} Appendix: MetaSymNet hyperparameter Settings.
\end{itemize}
\subsection*{\large Supplementary material on the experiment:}

\begin{itemize}
    \item \ref{AE} Appendix: Number of evaluations during inference
    \item \ref{AF} Appendix: Comparison of the full recovery rate.
    \item \ref{AG} Appendix: Test in NED (normalized edit distance).
    \item \ref{AH} Appendix: $R^2$-time Pareto plots of various algorithms on SRBench.
    \item \ref{MLP-SVR} Appendix: Comparison with traditional machine learning algorithms.
    \begin{itemize}
        \item \ref{AI1} Fit ability test
        \item \ref{AI2} Test for extrapolation
        \item \ref{AI3} Result complexity test
    \end{itemize}
    \item \ref{AK} Appendix: MetSymNet tests on AIFeynman dataset.
\end{itemize}
\subsection*{\large Other supplementary materials}
\begin{itemize}
    \item \ref{AJ} Appendix: Test data in detail.
    \item \ref{AL} Appendix: Computing resources
    
\end{itemize}

\newpage

\section{Appendix: Pseudocode for the MetaSymNet}
\label{AA}
Algorithm 1 describes the overall process of MetSymNet in detail. We first initialize a tree-like network structure, where the neurons of the internal nodes of the network are PANGU metafunctions and the neurons of the leaf nodes are Variable metafunctions. We first iterate $N_{DB}$times to optimize the amplitude parameter $W$ and bias parameter $B$ of each node. We then iterate $N_{ZW}$times to optimize the selection parameters $\mathbb{Z},\mathbb{W}$. We then use \textbf{ExtractingExpression}($\mathbb{Z}$, $\mathbb{D}$,$W$,$B$)\ref{algorithm2}  to evolve the neuron into a basic function, resulting in a concise expression. Then we can further optimize the parameters in the expression. Repeat the above process until $R^2$reaches the desired value,

\definecolor{mylightblue}{rgb}{0.48, 0.55, 0.95}

\begin{algorithm}[H]
  \caption{MetaSymNet}
  \label{pse1}
  \KwData{$X = [x_1,x_2,\dots,x_m]$; $y = [y_1,y_2,\dots,y_m]$; S =$[s_1,s_2,\dots,s_n]$;\\
  $W = [w_1,w_2,\dots,w_n]$; $B = [b_1,b_2,\dots,b_n]$;\\
  $\mathbb{Z} = [Z_1,Z_2,\dots,Z_n]$; $\mathbb{D} = [D_1,D_2,\dots,D_n]$.}
  \KwResult{Find an expression such that $y = f(X)$ }
    
  \textit{\textbf{Initialize the network structure}}\\
  \While{$R^2 \leq 0.9999$}{ 
    \Repeat{reaching predetermined accuracy}{
      \For{$j\leftarrow 1$ \KwTo $N_{WB}$}{
        \textbf{NumericalOptimization}($W$,$B$) \textcolor{gray}{\tcp{$N_{WB}$ denotes the number of optimization rounds for parameters $W$ and $B$.}}
      }
      \For{$j\leftarrow 1$ \KwTo $N_{ZD}$}{
        \textbf{NumericalOptimization}($\mathbb{Z}$, $\mathbb{D}$) \textcolor{gray}{\tcp{Optimizing the internal selection parameters $\mathbb{Z}, \mathbb{D}$ of PANGU metafunction and Variable metafunction.}}
      }
      \textbf{ExtractingExpression}($\mathbb{Z}$, $\mathbb{D}$, $W$, $B$)  \textcolor{gray}{\tcp{Based on the optimized parameters, a concise mathematical expression is extracted. algorithm \ref{algorithm2}}}
      \textbf{ConstantRefine}($W$, $B$) \textcolor{gray}{\tcp{Optimize constants in the extracted expression.}}
      \textbf{Obtain the formula}: $y_{pred} = F(x_1,x_2,\dots,x_n)$\\
      \textbf{Calculate} $R^2$\\
      \If{$R^2 > T$}{ 
        break \textcolor{gray}{\tcp{Terminate the program upon achieving expected $R^2$.}}
      }
    }     
  }
\end{algorithm}
\newpage
\section{Appendix: Pseudocode for how to evolve into expressions}
\label{AB}
\textbf{Algorithm 2} This pseudocode describes in detail the evolution process of PANGU meta-functions and Variable meta-functions in MetSymNet to basic functions and variables. For the PANGU meta-function, we first choose the index $ I$ corresponding to the maximum value in its function selection parameter $Z$. Then the corresponding symbol $ s$ is selected according to the index $I$. The principle of Variable meta-functions is similar to PANGU meta-functions, except that the target is a variable of type $[x_1,x_2,...,x_m]$.\\

\begin{algorithm}[H]
\caption{ExtractingExpression($\mathbb{Z}$, $\mathbb{D}$, $W$, $B$)}
\label{algorithm2}

\KwIn{$W = [w_1, w_2, \ldots, w_n]$; $B = [b_1, b_2, \ldots, b_n]$;\\
$\quad\mathbb{Z} = [Z_1, Z_2, \ldots, Z_n]$; $\mathbb{D} = [D_1, D_2, \ldots, D_w]$}
\KwOut{Expression $F = (x_1, \ldots, x_m)$}
$S \leftarrow [s_1, s_2, \ldots, s_n]$\\
$X \leftarrow [x_1, x_2, \ldots, x_m]$ \textcolor{gray}{\tcp{Candidate variables.}}
$Symbol \leftarrow [+, -, \times, \div, \sin, \ldots, x_1, \ldots, x_m]$ \textcolor{gray}{\tcp{Activation function library.}}
\For{$i \leftarrow len(S)$}{
  \uIf{$s_i$ is \texttt{PANGU\_metafunction}}{
    $I \leftarrow \text{arg max}(Z_{i})$ \textcolor{gray}{\tcp{The activation function is chosen according to the parameter $Z$.}}
    $s \leftarrow Symbol[I]$
  }
  \uElseIf{$s_i$ is \texttt{Variable\_metafunction}}{
    $v_i = \text{OUT}(s_i)$ \textcolor{gray}{\tcp{The output value of the node $s$ .}}
    $(I_l, I_r) \leftarrow \text{arg max-2}(D_{s})$ \textcolor{gray}{\tcp{Select the indices corresponding to the two variables with the largest selection parameters $D$.}}
    $(x_l, x_r) = (X[I_l], X[I_r])$ \textcolor{gray}{\tcp{Pick the two variables with the largest $D$.}}
    $O_{\text{out}} = [x_l + x_r, \ldots, \sin(x_l), \ldots, x_1, \ldots, x_n]$\textcolor{gray}{\tcp{Taking $x_l$,$x_r$ as input, all candidate operators are evaluated numerically.}}
    $I \leftarrow \text{arg min}(|O_{\text{out}} - v_i|)$\\
    $s \leftarrow Symbol[I]$\textcolor{gray}{\tcp{Select the symbol that is closest to the output value of node $s$ as the new activate function.}}
  }
}
\end{algorithm}
\section{Appendix: Backpropagation derivation (Use SGD)
}
\label{AC}
\textbf{Forward propagation : }
\begin{equation*}
\begin{aligned}
OUT_1 &= [sin(x), cos(x), ... ,exp(x)]*[\frac{e^{z_{11}}}{\sum_{i=1}^{n}{e^{z_{1i}}}}, \frac{e^{z_{12}}}{\sum_{i=1}^{n}{e^{z_{1i}}}}, ... ,\frac{e^{z_{1n}}}{\sum_{i=1}^{n}{e^{z_{1i}}}}]\\
    &=sin(x)*e_{11} + cos(x)*e_{12} + ... + exp(x)*e_{1n}\\
OUT_2 &= [sin(OUT_1), cos(OUT_1), ... ,exp(OUT_1)]*[\frac{e^{z_{21}}}{\sum_{i=1}^{n}{e^{z_{2i}}}}, \frac{e^{z_{22}}}{\sum_{i=1}^{n}{e^{z_{2i}}}}, ... ,\frac{e^{z_{2n}}}{\sum_{i=1}^{n}{e^{z_{2i}}}}]\\
&=sin(OUT_1)*e_{21} + cos(OUT_1)*e_{22} + ... + exp(OUT_1)*e_{2n}\\
Loss&=\frac{1}{2}\sum_{i=1}^{n}(Y_i-OUT_{2i})^2
\end{aligned}
\end{equation*}

\textbf{Back propagation : }
\begin{equation*}
\begin{aligned}
\frac{\partial Loss}{\partial OUT_2} &= OUT_2 - Y \\
\textcolor{red}{\frac{\partial Loss}{\partial z_{2i}}} &= \frac{\partial Loss}{\partial OUT_2} * \frac{\partial OUT_2}{\partial z_{2i}}\\ 
&=-\frac{\partial Loss}{\partial OUT_2}*[sin(OUT_1), cos(OUT_1), ... ,exp(OUT_1)]*e_{2i}(e_{21},...,e_{2i}-1,...,e_{2n})\\
&= (OUT_2 - Y)*[sin(OUT_1), cos(OUT_1), ... ,exp(OUT_1)]*e_{2i}(e_{21},...,e_{2i}-1,...,e_{2n})\\
\frac{\partial Loss}{\partial OUT_1} 
&= \frac{\partial Loss}{\partial OUT_2} * \frac{\partial OUT_2}{\partial OUT_1}\\
&=(OUT_2 - Y) * (cos(OUT_1)*e_{21} - sin(OUT_1)*e_{22} + ... + exp(OUT_1)*e_{2n})\\
\textcolor{red}{\frac{\partial Loss}{\partial z_{1i}}} &=\frac{\partial Loss}{\partial OUT_2} * \frac{\partial OUT_2}{\partial OUT_1}* \frac{\partial OUT_1}{\partial z_{1i}}\\
&=-\frac{\partial Loss}{\partial OUT_2} * \frac{\partial OUT_2}{\partial OUT_1}*[sin(x), cos(x), ... ,exp(x)]*e_{1i}(e_{11},...,e_{1i}-1,...,e_{1n})   
\end{aligned}
\end{equation*}
\textbf{Parameter updates}\\
\begin{equation*}
\begin{aligned}
z_{2i} &= z_{2i} -  \textcolor{red}{\alpha\frac{\partial Loss}{\partial z_{2i}}} \\
z_{1i} &= z_{1i} -  \textcolor{red}{\alpha\frac{\partial Loss}{\partial z_{1i}}}  \\
\end{aligned}
\end{equation*}

\section{Appendix: MetaSymNet hyperparameter Settings}
\label{AD}
MetaSymNet itself does not involve too many hyperparameters, and we list its hyperparameter Settings in Table \ref{tab-hyp}.
\begin{table*}[htbp]
\centering
{
\begin{tabular}{lcc}
\textbf{hyperparameters} & \textbf{Symbol}& \textbf{Numerical value}\\ 
\toprule
\textbf{Entropy loss regulation parameters} & $\lambda$& 0.2 \\
\textbf{[W,B] Optimization times}&$N_{WB}$ &  10\\
\textbf{[D,Z] Optimization times}&$N_{DZ}$ &  10\\
\textbf{$R^2$ termination threshold} & $\mathbf{T}$ & 0.9999\\
\textbf{Learning rate} &$\alpha$ & 0.01\\
\hline
\end{tabular} 
\caption{Hyperparameters of the Decoder of Transformer.}
\label{tab-hyp}
}
\end{table*}
\section{Appendix: Number of evaluations during inference}
\label{AE}
The number of evaluations represents the number of intermediate expressions experienced by the algorithm during inference.
For example, the expression $sin(x)$. MetaSymNet learning process is [$sin(x)+x$] $-->$ [$sin(x)+cos(x)$]$-->$[$sin(x+x)$]$-->$[$sin(x)$], which counts as 4 `evaluations'. The number of evaluations reflects the accuracy of the algorithm inference to a certain extent. The smaller the number of intermediate expressions experienced, the more accurate the algorithm grasped the inference direction. The number of evaluations of MetaSymNet and other baselines is shown in the following table \ref{tab-evaluation}. Our definition of `evaluation' will be different for each algorithm because of how it works. For DSO, for every batch (1000) of expressions we sample, we count as an `evaluation'. For TPSR and SPL, we count as an `evaluation' each time a full MCTS is performed. (Note: EQL and NeSymReS are not applicable for this evaluation).

From Table \ref{tab-evaluation}, we can see that the number of evaluations in the inference process of MetaSymNet is lower than other reinforcement learning-based algorithms. This shows the advantage of using numerical optimization in MetaSymNet.\\
\section{Appendix: Comparison of the full recovery rate}
\label{AF}
The full recovery rate of the algorithm is also an important index to evaluate the performance of the algorithm. The full recovery rate means that the final expression produced by the algorithm must be in the same form as the original \\
\begin{table}
\center
\vspace{-0.2cm}

\resizebox{8cm}{!}{
\begin{tabular}{ccccc}
 \toprule
 Dataset   & MetaSymNet & DSO     & TPSR    & SPL \\ 
 \toprule
 Nguyen-1  & 7.0        & 7.9     & 10.2  & 14.5   \\
 Nguyen-2  & 10.4       & 10.5    & 14.3  & 18.2  \\
 Nguyen-3  & 12.5       & 16.3    & 18.5  & 23.7  \\
 Nguyen-4  & 17.2       & 20.6    & 24.8  & 25.3  \\
 Nguyen-5  & 19.4       & 28.0    & 25.2  & 27.2 \\
 Nguyen-6  & 14.2       & 20.7    & 22.6  & 36.4  \\
 Nguyen-7  & 18.0       & 24.3    & 26.9  & 34.0 \\
 Nguyen-8  & 9.5        & 1.1     & 6.6   & 7.9   \\ 
 Nguyen-9  & 19.0       & 28.5    & 24.4  & 37.5  \\ 
 Nguyen-10 & 18.2       & 29.4    & 32.6  & 36.2 \\
 Nguyen-11 & 16.4       & 19.8    & 26.7  & 33.0  \\
 Nguyen-12 & 26.0       & 28.5    & 27.2  & 36.3 \\
  \cline{2-5}
 Average   & 15.6       & 19.6    & 21.7  & 27.5  \\
\toprule
\end{tabular}
}
\caption{The table shows the comparison of the number of evaluations during the search of the four algorithms.}
\label{tab-evaluation}
\end{table}

\begin{figure*}
\centering
\hspace*{-0.6cm}
\includegraphics[width=180mm]{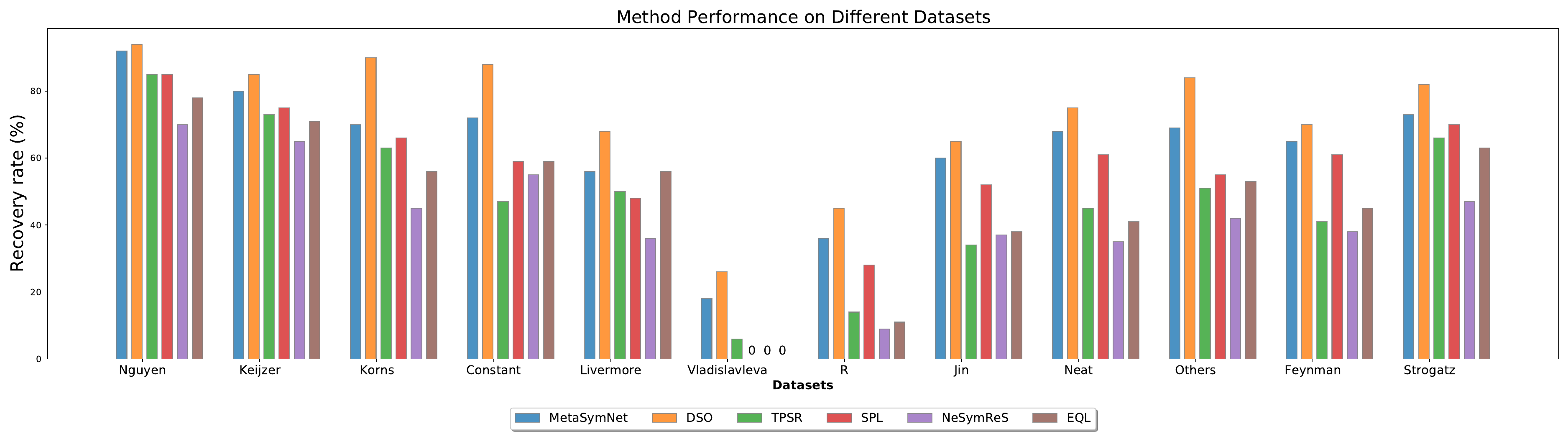}
\caption{The figure shows the full recovery rate of each algorithm on different datasets. Note: we ran each expression 50 times.} 
\label{fig-recovery-rate}
\end{figure*}
expression ($R^2 \approx 1.0$), and not an approximate expression. This requires a manual one-to-one comparison. We ran each expression ten times and recorded how many times the expression was completely recovered.
Figure \ref{fig-recovery-rate} shows the full recovery rate of each algorithm on different datasets. From the figure, we can see that MetaSymNet is slightly lower than DSO in the full recovery rate as a whole, but on some datasets with more complex expressions (e.g. Vladislavleva, R, Feynman), MetaSymNet's recovery rate is better than DSO. This shows that MetaSymNet has great potential to deal with more complex problems. Because MetaSymNet can also be viewed as a neural network, it inherits the ability of ordinary neural networks to handle complex problems.
\section{Appendix: Test in NED (normalized edit distance)}
\label{AG}
In order to better test the performance of MetaSymNet and other baselines, we also use normalized edit distances (NED) proposed in article  \cite{matsubara2022rethinking} to test each baseline. The NED expression is as follows Eq\ref{e_NED}.
\begin{equation}
\label{e_NED}
NED(f_{pred},f_{true})=min(1,\frac{NED(f_{pred},f_{true})}{|f_{true}|})
\end{equation}
$f_{pred}$ and $f_{true}$ are the estimated and true equation trees, respectively. $NED(f_{pred}, ftrue)$ is the edit distance between $f_{pred}$ and $f_{true}$. $|f_{true}|$ denotes the number of tree nodes composing equation $f_{true}$. We note that this metric is meant to capture the symbolic similarity between the predicted and true equations, so the constant values themselves are not important, so we only consider operators and not all constants (or constant placeholders).
The data set used for testing is the SRSD data set. The result is as follows table \ref{table:baseline_results}, and  table\ref{table:baseline_results_w_dummy_vars}:
\begin{table*}[t]
    
    \def\arraystretch{1.2}
    \small
    \begin{center}
        \bgroup
        \setlength{\tabcolsep}{0.4em}
        \begin{tabular}{c|l|rrrrrr} 
            \toprule
            \multirow{2}{*}{\bf Metric} & \multicolumn{1}{c|}{\multirow{2}{*}{\bf Group}} & \multicolumn{6}{c}{\bf SRSD-Feynman} \\
            \cline{3-8}
            & & \multicolumn{1}{c}{\bf MetaSymNet} & \multicolumn{1}{c}{\bf DSO} & \multicolumn{1}{c}{\bf TPSR} & \multicolumn{1}{c}{\bf SPL} & \multicolumn{1}{c}{\bf NeSymReS} & \multicolumn{1}{c}{\bf EQL}  \\
            \midrule
            \multirow{3}{*}{\rotatebox{90}{NED}} & Easy & 0.533&	\textbf{0.448}&	0.630&	0.793 &  0.825 & 0.782  \\ 
            & Medium & \textbf{0.615}&	0.691&	0.792&	0.848 & 0.891 & 0.823  \\ 
            & Hard & \textbf{0.641}&	0.744&	0.915&0.952 & 0.979 & 0.872 \\
            \bottomrule
        \end{tabular}
        \egroup
        \caption{Baseline results for NED (normalized edit distance).}
    \label{table:baseline_results}
    \end{center}    
    
    \def\arraystretch{1.2}
    \small
    \begin{center}
        \bgroup
        \setlength{\tabcolsep}{0.4em}
        \begin{tabular}{c|l|rrrrrr} 
            \toprule
            \multirow{2}{*}{\bf Metric} & \multicolumn{1}{c|}{\multirow{2}{*}{\bf Group}} & \multicolumn{6}{c}{\bf SRSD-Feynman} \\
            \cline{3-8}
            & & \multicolumn{1}{c}{\bf MetaSymNet} & \multicolumn{1}{c}{\bf DSO} & \multicolumn{1}{c}{\bf TPSR} & \multicolumn{1}{c}{\bf SPL} & \multicolumn{1}{c}{\bf NeSymReS} & \multicolumn{1}{c}{\bf EQL}  \\
            \midrule
            \multirow{3}{*}{\rotatebox{90}{NED}} & Easy & 0.703&	\textbf{0.681}&	0.863&	0.906& 0.951 & 0.899  \\ 
            & Medium & \textbf{0.660}&	0.751&	0.962&	0.978& 0.981 & 0.863  \\ 
            & Hard & \textbf{0.693}&	0.746&	1.000&	1.000
 & 1.000 & 0.977 \\
            \bottomrule
        \end{tabular}
        \egroup
        \caption{Baseline results for NED (normalized edit distance).}
        \label{table:baseline_results_w_dummy_vars}
    \end{center}
\end{table*}
\section{Appendix: $R^2$-time Pareto plots of various algorithms on SRBench}
\label{AH}
\begin{figure*}
\centering
\includegraphics[width=140mm]{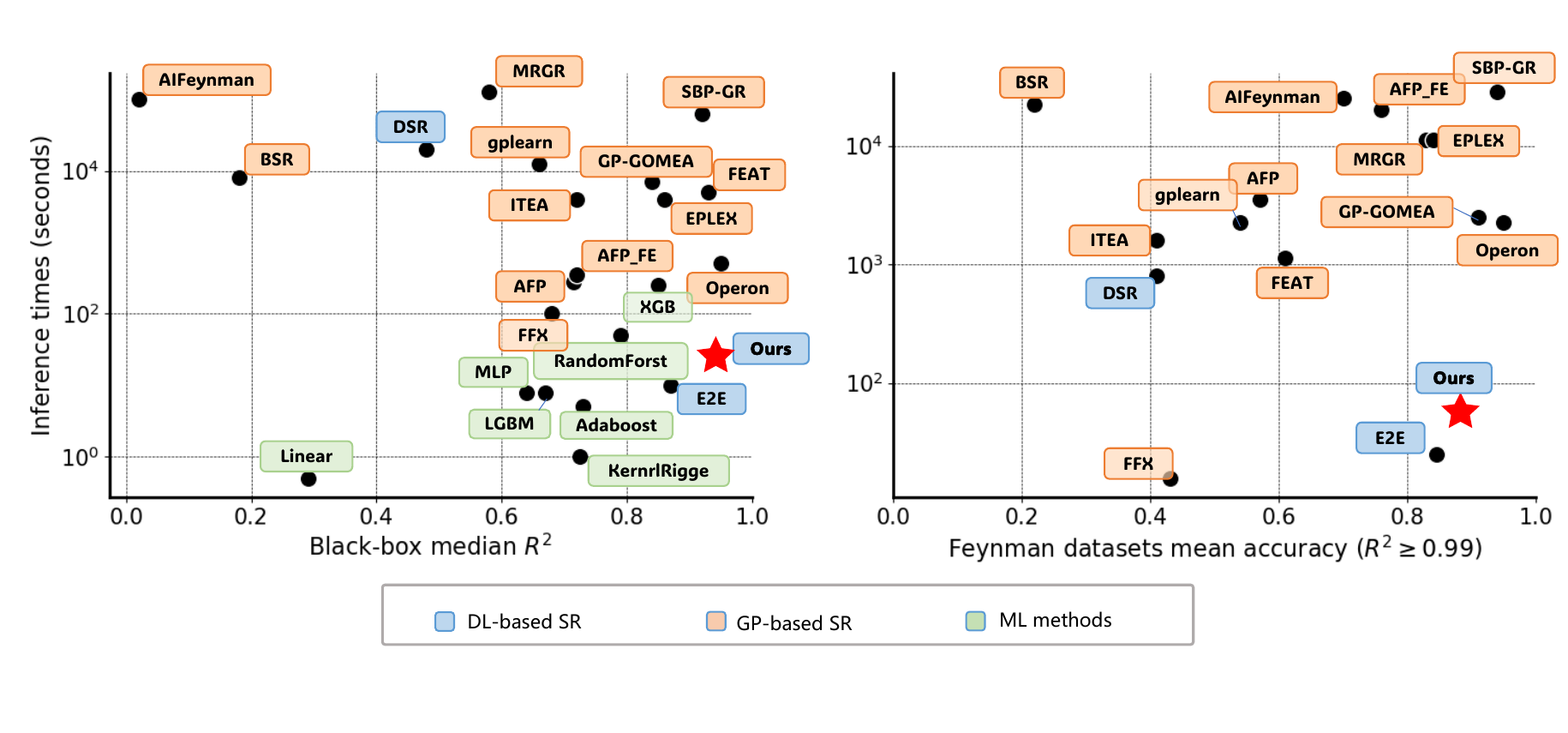}

\definecolor{lightblue}{RGB}{150,216,255}
\caption{MetaSymNet outperforms many baselines. Compared with SOTA (GP-based method), MetasymNet has significantly improved inference speed. We compared MetaSymNet's average test performance and inference time to baselines provided by the SRbench benchmark on the Feynman SR problem and black-box regression problem. Color is used to distinguish between three model families: \textcolor{lightblue}{deep-learning-based SR}, \textcolor{orange}{genetic programming-based SR}, and \textcolor{green}{classical machine learning approaches} {which do not provide symbolic solutions}.} 
\label{fig-SRBench}
\end{figure*}

This section tests the $R^2$-time Pareto plots of MetaSymNet and multiple baselines on datasets Black-box and Feynman. In fig.\ref{fig-SRBench}, the closer the coordinates of the algorithm are to the bottom-right corner, the better the overall performance. As we can see, MetaSymNet(Red star) is very close to the bottom-right corner of the figure, which has a good overall performance.
\section{Appendix: Comparison with traditional machine learning algorithms}
\label{MLP-SVR}
To test the comprehensive performance of MetaSymNet compared with traditional neural Network (MLP)  \citep{bpnn} and Support Vector Machine (SVM)  \citep{svr}. We conduct a series of test experiments on the performance of MetaSymNet, MPL, and SVM on the Nguyen dataset. The experiments mainly include the test of \textbf{fitting ability, extrapolation test, and network structure complexity test.}
\subsection{Fit ability test}
\label{AI1}
As the name suggests, fit ability tests how well three algorithms fit the data. In the test, we sample $x_i$ on the interval [-2,2], and then use three algorithms to fit the same data points, using $R^2$ as the evaluation metric. The experimental results are shown in Table \ref{tab2} [-2,2], from which it can be seen that MetaSymNet slightly outperforms MLP and SVR in terms of fitting ability.
\subsection{Test for extrapolation}
\label{AI2}
Extrapolation is the ability to test how well the model fits outside of the training data. Let's say we sample the variable $x$ in the interval [-2, 2] and use this data to train a model. The question arises whether a model that exhibits a good fit for data within the interval [-2, 2] will maintain equivalent performance when extended to the interval [-5, 5]. 

\begin{figure*}[htp]
    \centering
    \vspace{-1.6cm}
    \setlength{\belowcaptionskip}{-0.3cm} 
       \subfloat[]{
       \includegraphics[width=0.47\linewidth]{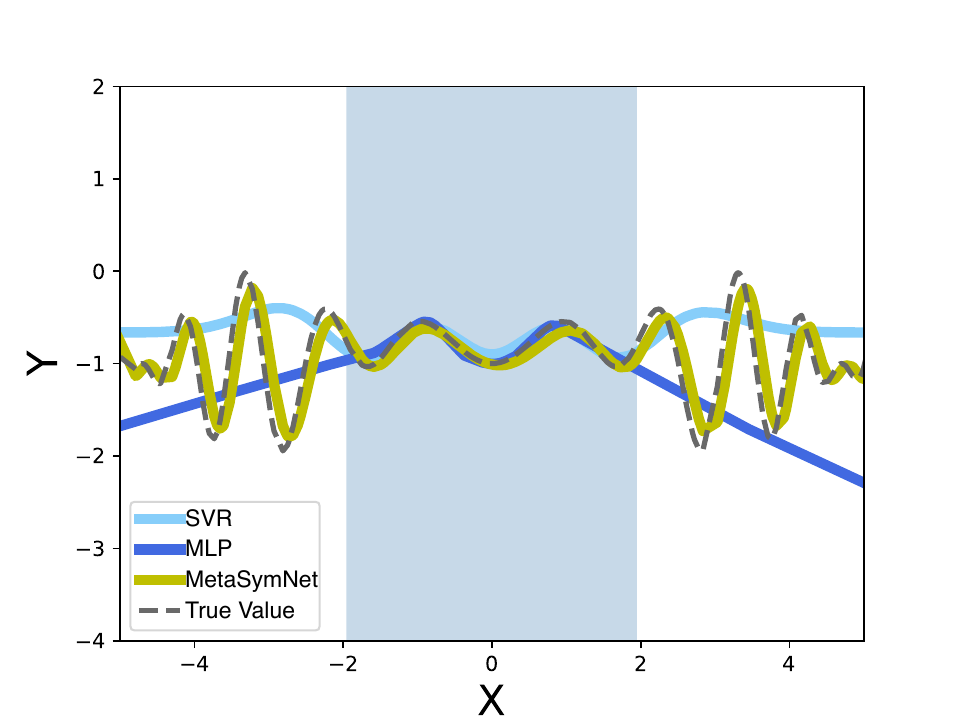} \label{fig3a}}
        \subfloat[]{
        \includegraphics[width=0.48 \linewidth]{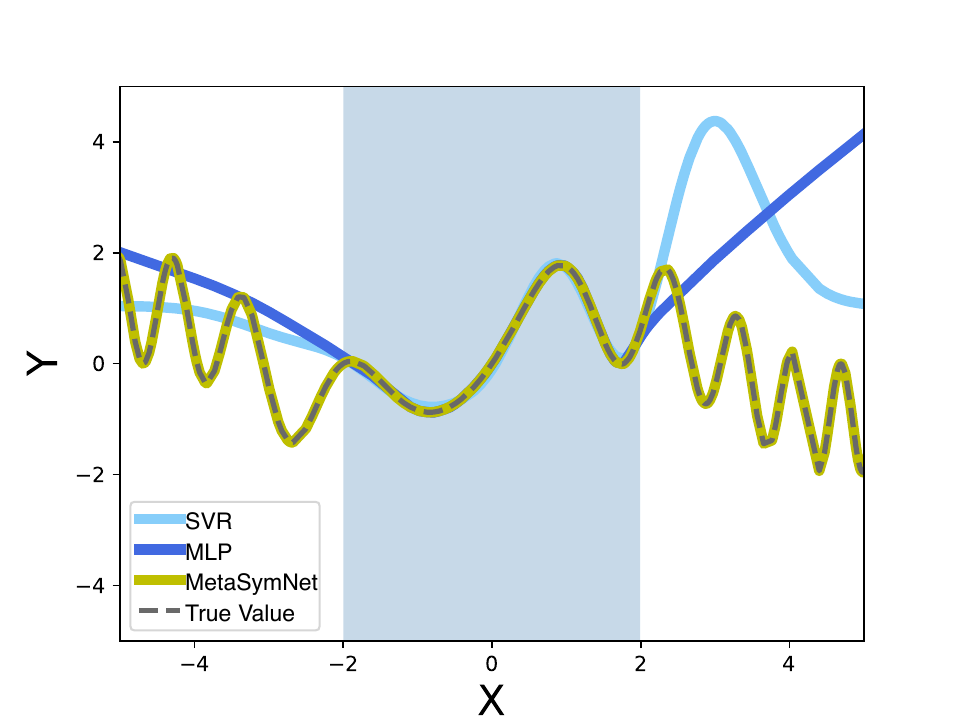}\label{fig3b}}   
	  \caption{
   Performance comparison of MetaSymNet, MLP, and SVR in terms of extrapolation ability. (Nguyen-5 : $sin(x^2)cos(x)-1$; Nguyen-6 : $sin(x) + sin(x + x^2)$).    
  }
\label{fig3} 
\end{figure*}
To test this, we sample $x$ on the interval [-2,2] and train MetaSymNet, MLP, and SVR on the sampled data. The three models, after training, underwent testing within the intervals [-2, 2] and [-5, 5], respectively. Table \ref{tab2} shows the performance of MetaSymNet, MLP, and SVR. We can see that MetaSymNet and both baselines perform well in the range [-2,2]. However, upon expanding the interval to [-5, 5], MetaSymNet substantially surpasses the performance of the two baseline algorithms. Because the activation functions of MetaSymNet are more diverse, the MetaSymNet has a stronger representation ability. Therefore, compared with MLP, it is easier to obtain results with stronger extrapolation. The picture of the extrapolation effect is shown in Figure \ref{fig3b}
\subsection{Result complexity test}
\label{AI3}
The network structure of the MLP is fixed, and it can not adjust network structure dynamically according to the tasks. The design may readily induce a phenomenon characterized by redundant network architecture and superfluous parameters. MetaSymNet's network structure is dynamically adjusted according to the tasks, which can greatly improve the network structure and parameter redundancy. To test the structural optimization ability of MetaSymNet, we compare the resulting complexity of MLP(using an iterative pruning strategy   \citep{prun}) and MetaSymNet through the Nguyen dataset. Specifically, we compare the number of neurons and parameters used by the two algorithms when $R^2$ reaches 0.99. The detailed statistical results are shown in Table \ref{tab3}. As can be seen from the table, MetaSymNet has a simpler architecture and a smaller number of parameters at the same level of $R^2$, And the more complex the problem, the more obvious this advantage will be.
\begin{table*}[ht]
\center
\resizebox{18cm}{!}{
\begin{tabular}{cccccccc}

\toprule
Benchmark&Expression & \multicolumn{2}{c}{MetaSymNet}& \multicolumn{2}{c}{MLP}& \multicolumn{2}{c}{SVR}\\
\toprule
   &    & [-2,2] & [-5,5]  & [-2,2]& [-5,5]  & [-2,2]&[-5,5] \\ 
\cmidrule(lr){1-2}
\cmidrule(lr){3-4}
\cmidrule(lr){5-6}
\cmidrule(lr){7-8}
Nguyen-1 & $x^3+x^2+x$ & $\textbf{1.0}$ & $\textbf{1.0}$ & $0.9999$ &$ -32.47$ & $0.9999$ & $-59.33$ \\
Nguyen-2 & $x^4+x^3+x^2+x$ & $\textbf{1.0}$ &$ \textbf{1.0} $& $0.9999$ & $-121.36$ & $0.9998$ & $-425.28$ \\
Nguyen-3 & $x^5+x^4+x^3+x^2+x$ & $\textbf{0.9999}$ & $\textbf{0.9725}$ & $\textbf{0.9999}$ & $-602.86$ & $0.9942 $& $-410.12$ \\
Nguyen-4 & $x^6+x^5+x^4+x^3+x^2+x$ & $\textbf{0.9999}$ &$\textbf{0.9382}$ & $0.9982$ &$-3002.18$ & $0.9847$& $-4908.16$  \\
Nguyen-5 & $\sin(x^2)\cos(x)-1$ & $\textbf{0.9999} $& $\textbf{0.2344}$ & $0.9916$& $-12.46$ & $0.5322$ & $-27.36$\\
Nguyen-6 & $\sin(x)+\sin(x+x^2)$ & $\textbf{1.0}$ & $\textbf{1.0}$ & $0.9992$ & $-10.15$ & $0.9918$ & $-3.34$ \\
Nguyen-7 & $\log(x+1)+\log(x^2+1)$ & $0.9938$ & $\textbf{0.9462}$ & $\textbf{0.9999}$ &$0.56$ & $0.9817$ & $-60.15$ \\
Nguyen-8 &$ \sqrt{x}$ & $\textbf{1.0}$ & $\textbf{1.0}$ & $0.9998$ & $0.69$& $0.9493$ &$-48.42$\\
Nguyen-9 & $\sin(x)+\sin(y^2)$ & $\textbf{1.0}$ & $\textbf{1.0}$ & $0.9999$ & $-2.21$ & $0.9444$ & $-52.39$\\
Nguyen-10 &$ 2\sin(x)\cos(y)$ & $\textbf{1.0}$ & $\textbf{1.0}$ & $0.9949 $&$ -18.56$ & $0.9863$& $-2.95 $\\
Nguyen-11 &$ x^y$ & $\textbf{0.9999}$ &$\textbf{0.9921}$ & $\textbf{0.9999}$ &$-137892$ & $0.9924$ & $-901002$ \\
Nguyen-12 &$ x^4-x^3+\frac{1}{2}y^2-y$ & $\textbf{0.9972} $& $\textbf{-1.818}$ & $0.9928$ & $-610.21$& $0.9904$ & $-7822.80$ \\ 
 \cline{3-8}
 & \multicolumn{1}{r}{Average} & $\textbf{\textbf{0.9999}}$ & $\textbf{0.6888}$& $0.9980$ & $-11858.60$ & $0.9456$& $-76235.19 $ \\
 \toprule
\end{tabular}
}
\caption{Comparison of extrapolation and $R^2$ between MetaSymNet, MLP, and SVR.
. \label{tab2}}
\end{table*}
\begin{table}
\center
\vspace{-0.2cm}
\resizebox{9cm}{!}{
\begin{tabular}{ccccc}
Benchmark& \multicolumn{2}{c}{MetaSymNet}& \multicolumn{2}{c}{MLP}\\ \toprule
      & Nodes & Parameters  & Nodes& Parameters   \\ 
      \cmidrule(lr){2-3}
      \cmidrule(lr){4-5}
Nguyen-1 & 5  & 22 & 15 &  70  \\
Nguyen-2 & 9 & 38  & 17 & 94 \\
Nguyen-3 & 14 & 58 & 18 & 106 \\
Nguyen-4 & 20 & 82 & 25 &185 \\
Nguyen-5 & 5 & 16 & 25& 185 \\
Nguyen-6 & 5 & 18 & 8 & 22  \\
Nguyen-7 & 4 & 16 & 12 &34 \\
Nguyen-8 & 1 & 4 & 16 & 82\\
Nguyen-9 & 3 & 10 & 40 & 360\\
Nguyen-10 & 4 & 14 & 130 & 3190 \\
Nguyen-11 & 3 & 6 & 16 &88  \\
Nguyen-12 & 10 & 40 & 16 & 96 \\ 
 \cline{1-5}
Average & \textbf{6.75} & \textbf{27} & 28.16 & 376  \\
\toprule
\end{tabular}
}
\caption{Comparison of the number of nodes and parameters used by MetaSymNet and MLP when the $R^2$ is greater than 0.99.\label{tab3}}
\end{table}

\section{Appendix: Test data in detail}
\label{AJ}
\cref{a-tab1,a-tab2,a-tab3} shows in detail the expression forms of the data set used in the experiment, as well as the sampling range and sampling number. Some specific presentation rules are described below
\begin{itemize}
\item The variables contained in the regression task are represented as [$x_1,x_2,...,x_n$].
\item $U(a,b,c)$ signifies $c$ random points uniformly sampled between $a$ and $b$ for each input variable. Different random seeds are used for training and testing datasets.
\item $E(a,b,c)$ indicates $c$ points evenly spaced between $a$ and $b$ for each input variable. 

\end{itemize}

\begin{table*}[htbp]
\centering
\begin{scriptsize}
\begin{tabular}{ccccc}
\toprule[1.45pt]
\toprule
Name & Expression & Dataset  \\ \hline
Nguyen-1 & $x_1^3+x_1^2+x_1$&U$(-1, 1, 20)$\\
Nguyen-2 & $x_1^4+x_1^3+x_1^2+x_1$ & U$(-1, 1, 20)$ \\
Nguyen-3 & $x_1^5+x_1^4+x_1^3+x_1^2+x_1$ & U$(-1, 1, 20)$ \\
Nguyen-4 & $x_1^6+x_1^5+x_1^4+x_1^3+x_1^2+x_1$ & U$(-1, 1, 20)$  \\
Nguyen-5 & $\sin(x_1^2)\cos(x)-1$ & U$(-1, 1, 20)$  \\
Nguyen-6 & $\sin(x_1)+\sin(x_1+x_1^2)$ & U$(-1, 1, 20)$  \\
Nguyen-7 & $\log(x_1+1)+\log(x_1^2+1)$ & U$(0, 2, 20)$  \\
Nguyen-8 & $\sqrt{x}$ & U$(0, 4, 20)$  \\
Nguyen-9 & $\sin(x)+\sin(x_2^2)$ & U$(0, 1, 20)$ \\
Nguyen-10 & $2\sin(x)\cos(x_2)$ & U$(0, 1, 20)$ \\
Nguyen-11 & $x_1^{x_2}$ & U$(0, 1, 20)$  \\
Nguyen-12 & $x_1^4-x_1^3+\frac{1}{2}x_2^2-x_2$ & U$(0, 1, 20)$ \\
\toprule
Nguyen-2$'$ & $4x_1^4+3x_1^3+2x_1^2+x$ & U$(-1, 1, 20)$  \\
Nguyen-5$'$ & $\sin(x_1^2)\cos(x)-2$ & U$(-1, 1, 20)$  \\
Nguyen-8$'$ & $\sqrt[3]{x}$ & U$(0, 4, 20)$ \\
Nguyen-8$''$ & $\sqrt[3]{x_1^2}$ & U$(0, 4, 20)$ \\
\toprule
Nguyen-1\textsuperscript{c} & $3.39x_1^3+2.12x_1^2+1.78x$ & U$(-1, 1, 20)$ \\
Nguyen-5\textsuperscript{c} & $\sin(x_1^2)\cos(x)-0.75$ & $U(-1, 1, 20)$  \\
Nguyen-7\textsuperscript{c} & $\log(x+1.4)+\log(x_1^2+1.3)$ & U$(0, 2, 20)$ \\
Nguyen-8\textsuperscript{c} & $\sqrt{1.23 x}$ & U$(0, 4, 20)$  \\
Nguyen-10\textsuperscript{c} & $\sin(1.5x)\cos(0.5x_2)$ & U$(0, 1, 20)$  \\
\toprule
Korns-1 & $1.57+24.3*x_1^4$ & U$(-1, 1, 20)$  \\
Korns-2 & $0.23+14.2\frac{(x_4+x_1)}{(3x_2)}$ & U$(-1, 1, 20)$  \\
Korns-3 & $4.9\frac{(x_2-x_1+\frac{x_1}{x_3}}{(3x_3))}-5.41$ & U$(-1, 1, 20)$ \\
Korns-4 & $0.13sin(x_1)-2.3$ & U$(-1, 1, 20)$  \\
Korns-5 & $3+2.13log(|x_5|)$ & U$(-1, 1, 20)$  \\
Korns-6 & $1.3+0.13\sqrt{|x_1|}$ & U$(-1, 1, 20)$  \\
Korns-7 & $2.1(1-e^{-0.55x_1})$ & U$(-1,1 , 20)$  \\
Korns-8 & $6.87+11\sqrt{|7.23 x_1 x_4 x_5|}$ & U$(-1, 1, 20)$ \\
Korns-9 & $12\sqrt{|4.2x_1x_2x_2|}$ & U$(-1, 1, 20)$ \\
Korns-10 & $0.81+24.3\frac{2x_{1}+3x_2^2}{4x_3^3+5x_4^4}$ & U$(-1, 1, 20)$  \\
Korns-11 & $6.87+11cos(7.23x_1^3)$ & U$(-1, 1, 20)$  \\
Korns-12 & $2-2.1cos(9.8x_1^3)sin(1.3x_5)$ & U$(-1, 1, 20)$  \\ 
Korns-13 & $32.0-3.0\frac{tan(x_1)}{tan(x_2)}\frac{tan(x_3)}{tan(x_4)}$ & U$(-1, 1, 20)$ \\
Korns-14 & $22.0-(4.2cos(x_1)-tan(x_2))\frac{tanh(x_3)}{sin(x_4)}$ & U$(-1, 1, 20)$  \\
Korns-15 & $12.0-\frac{6.0tan(x_1)}{e^{x_2}}(log(x_3)-tan(x_4))))$ & U$(-1, 1, 20)$  \\ 
\toprule
Jin-1 & $2.5 x_1^4-1.3 x_1^3 +0.5 x_2^2 - 1.7x_2$ & U$(-3, 3, 100)$ \\
Jin-2 & $8.0 x_1^2 + 8.0 x_2^3 - 15.0$ & U$(-3, 3, 100)$  \\
Jin-3 & $0.2 x_{1}^{3} + 0.5 x_{2}^{3} - 1.2 x_2 - 0.5 x_{1}$ & U$(-3, 3, 100)$  \\    
Jin-4 & $1.5 \exp{x} + 5.0 cos(x_2)$ & U$(-3, 3, 100)$\\
Jin-5 & $6.0 sin(x_1) cos(x_2)$ & U$(-3, 3, 100)$ \\
Jin-6 & $1.35 x_1 x_2 + 5.5 sin((x_1 - 1.0)(x_2 - 1.0))$ & U$(-3, 3, 100)$ \\ 
\toprule
\newline
\end{tabular}
\end{scriptsize}
\caption{ Specific formula form and value range of the three data sets Nguyen, Korns, and Jin. 
}
\label{a-tab1}
\end{table*}

\begin{table*}[htpb]
\centering

\begin{scriptsize}
\begin{tabular}{ccccc}
\toprule[1.45pt]
\toprule
Name & Expression & Dataset \\
\hline
Neat-1 & $x_1^4+x_1^3+x_1^2+x$ & U$(-1, 1, 20)$  \\
Neat-2 & $x_1^5+x_1^4+x_1^3+x_1^2+x$ & U$(-1, 1, 20)$ \\
Neat-3 & $\sin(x_1^2)\cos(x)-1$ & U$(-1, 1, 20)$ \\
Neat-4 & $\log(x+1)+\log(x_1^2+1)$ & U$(0, 2, 20)$  \\
Neat-5 & $2\sin(x)\cos(x_2)$ & U$(-1, 1, 100)$  \\
Neat-6 & $\sum_{k=1}^x \frac{1}{k} $ & E$(1, 50, 50)$  \\
Neat-7 & $2 - 2.1\cos(9.8x_1)\sin(1.3x_2)$ & E$(-50, 50, 10^5)$ \\
Neat-8 & $\frac{e^{-(x_1)^2}}{1.2 + (x_2-2.5)^2}$ & U$(0.3, 4, 100)$  \\
Neat-9 & $\frac{1}{1+x_1^{-4}} + \frac{1}{1+x_2^{-4}}$ & E$(-5, 5, 21)$ \\
\toprule
Keijzer-1 & $0.3x_1sin(2\pi x_1)$ & U$(-1, 1, 20)$  \\
Keijzer-2 & $2.0x_1sin(0.5\pi x_1)$ & U$(-1, 1, 20)$  \\
Keijzer-3 & $0.92x_1sin(2.41\pi x_1)$ & U$(-1, 1, 20)$ \\
Keijzer-4 & $x_1^3e^{-x_1}cos(x_1)sin(x_1)sin(x_1)^{2}cos(x_1)-1$ & U$(-1, 1, 20)$ \\
Keijzer-5 & $3+2.13log(|x_5|)$ & U$(-1, 1, 20)$\\

Keijzer-6 & $\frac{x1(x1+1)}{2}$& U$(-1, 1, 20)$ \\
Keijzer-7 & $log(x_1)$ & U$(0,1 , 20)$ \\
Keijzer-8 & $\sqrt{(x_1)}$ & U$(0, 1, 20)$  \\
Keijzer-9 & $log(x_1+\sqrt{x_1^2}+1)$ & U$(-1, 1, 20)$ \\
Keijzer-10 & $x_{1}^{x_2}$ & U$(-1, 1, 20)$  \\
Keijzer-11 & $x_1x_2+sin((x_1-1)(x_2-1))$ & U$(-1, 1, 20)$  \\
Keijzer-12 & $x_1^4-x_1^3+\frac{x_2^2}{2}-x_2$ & U$(-1, 1, 20)$  \\ 
Keijzer-13 & $6sin(x_1)cos(x_2)$ & U$(-1, 1, 20)$  \\
Keijzer-14 & $\frac{8}{2+x_1^2 + x_2^2}$ & U$(-1, 1, 20)$ \\
Keijzer-15 & $\frac{x_1^3}{5}+\frac{x_2^3}{2}-x_2-x_1$ & U$(-1, 1, 20)$ \\ 

\toprule
Livermore-1 & $\frac{1}{3}+x_1+sin(x_1^2))$ & U$(-3, 3, 100)$  \\
Livermore-2 & $sin(x_1^2)*cos(x1)-2$ & U$(-3, 3, 100)$  \\
Livermore-3 & $sin(x_1^3)*cos(x_1^2))-1$ & U$(-3, 3, 100)$  \\
Livermore-4 & $log(x_1+1)+log(x_1^2+1)+log(x_1)$ & U$(-3, 3, 100)$ \\ 
Livermore-5 & $x_1^4-x_1^3+x_2^2-x_2$ & U$(-3, 3, 100)$  \\
Livermore-6 & $4x_1^4+3x_1^3+2x_1^2+x_1$ & U$(-3, 3, 100)$ \\ 
Livermore-7 & $\frac{(exp(x1)-exp(-x_1)}{2})$ & U$(-1, 1, 100)$ \\ 
Livermore-8 & $\frac{(exp(x1)+exp(-x1)}{3}$ & U$(-3, 3, 100)$ \\
Livermore-9 & $x_1^9+x_1^8+x_1^7+x_1^6+x_1^5+x_1^4+x_1^3+x_1^2+x_1$ & U$(-1, 1, 100)$  \\
Livermore-10 & $6*sin(x_1)cos(x_2)$ & U$(-3, 3, 100)$  \\
Livermore-11 & $\frac{x_1^2 x_2^2}{(x_1+x_2)}$ & U$(-3, 3, 100)$ \\
Livermore-12 & $\frac{x_1^5}{x_2^3}$ & U$(-3, 3, 100)$  \\
Livermore-13 & $x_1^{\frac{1}{3}}$ & U$(-3, 3, 100)$  \\
Livermore-14 & $x_1^3+x_1^2+x_1+sin(x_1)+sin(x_2^2)$ & U$(-1, 1, 100)$ \\ 
Livermore-15 & $x_1^\frac{1}{5}$ & U$(-3, 3, 100)$  \\
Livermore-16 & $x_1^{\frac{2}{3}}$ & U$(-3, 3, 100)$  \\  
Livermore-17 & $4sin(x_1)cos(x_2)$ & U$(-3, 3, 100)$  \\
Livermore-18 & $sin(x_1^2)*cos(x_1)-5$ & U$(-3, 3, 100)$  \\
Livermore-19 & $x_1^5+x_1^4+x_1^2 + x_1$ & U$(-3, 3, 100)$  \\
Livermore-20 & $e^{(-x_1^2)}$ & U$(-3, 3, 100)$  \\
Livermore-21 & $x_1^8+x_1^7+x_1^6+x_1^5+x_1^4+x_1^3+x_1^2+x_1$& U$(-1, 1, 20)$ \\
Livermore-22 & $e^{(-0.5x_1^2)}$ & U$(-3, 3, 100)$  \\
\toprule
\newline
\end{tabular}
\end{scriptsize}
\caption{
Specific formula form and value range of the three data sets neat, Keijzer, and Livermore.
}
\label{a-tab2}
\end{table*}

\begin{table*}[htpb]
\centering

\begin{scriptsize}
\begin{tabular}{ccccc}
\toprule[1.45pt]
\toprule
Name & Expression & Dataset \\
\toprule
Vladislavleva-1 & $\frac{(e^{-(x1-1)^2})}{(1.2+(x2-2.5)^2))}$ & U$(-1, 1, 20)$ \\
Vladislavleva-2 & $e^{-x_1}x_1^3cos(x_1)sin(x_1)(cos(x_1)sin(x_1)^2-1)$ & U$(-1, 1, 20)$ \\

Vladislavleva-3 & $e^{-x_1}x_1^3cos(x_1)sin(x_1)(cos(x_1)sin(x_1)^2-1)(x_2-5)$ & U$(-1, 1, 20)$ \\
Vladislavleva-4 & $\frac{10}{5+(x1-3)^2+(x_2-3)^2+(x_3-3)^2+(x_4-3)^2+(x_5-3)^2}$ & U$(0, 2, 20)$ \\
Vladislavleva-5 & $30(x_1-1)\frac{x_3-1}{(x_1-10)}x_2^2$ & U$(-1, 1, 100)$ \\
Vladislavleva-6 & $6sin(x_1)cos(x_2)$ & E$(1, 50, 50)$ \\
Vladislavleva-7 & $2 - 2.1\cos(9.8x)\sin(1.3x_2)$ & E$(-50, 50, 10^5)$ \\
Vladislavleva-8 & $\frac{e^{-(x-1)^2}}{1.2 + (x_2-2.5)^2}$ & U$(0.3, 4, 100)$  \\
\toprule
Test-2 & $3.14x_1^2$ & U$(-1, 1, 20)$ \\
Const-Test-1 & $5x_1^2$ & U$(-1, 1, 20)$ \\
GrammarVAE-1 & $1/3+x1+sin(x_1^2))$ & U$(-1, 1, 20)$ \\
Sine & $sin(x_1)+sin(x_1+x_1^2))$ & U$(-1, 1, 20)$ \\
Nonic & $x_1^9+x_1^8+x_1^7+x_1^6+x_1^5+x_1^4+x_1^3+x_1^2+x_1$ & U$(-1, 1, 100)$  \\
Pagie-1 & $\frac{1}{1+x_1^{-4}+\frac{1}{1+x2^{-4}}} $ & E$(1, 50, 50)$  \\
Meier-3 & $\frac{x_1^2  x_2^2}{(x_1+x_2)}$ & E$(-50, 50, 10^5)$ \\
Meier-4 & $\frac{x_1^5}{x_2^3}$ & $U(0.3, 4, 100)$  \\
Poly-10 & $x_1x_2+x_3x4+x_5x_6+x_1x_7x_9+x_3x_6x_{10}$ & E$(-1, 1, 100)$ \\
\toprule
Constant-1 & $3.39*x_1^3+2.12*x_1^2+1.78*x_1$&$U(-4, 4, 100)$\\
Constant-2 & $sin(x_1^2)*cos(x_1)-0.75$&$U(-4, 4, 100)$\\
Constant-3 & $sin(1.5*x_1)*cos(0.5*x_2)$&$U(0.1, 4, 100)$\\
Constant-4 & $2.7*x_1^{x_2}$&$U(0.3, 4, 100)$\\
Constant-5 & $sqrt(1.23*x_1)$&$U(0.1, 4, 100)$\\
Constant-6 & $x_1^{0.426}$&$U(0.0, 4, 100)$\\
Constant-7 & $2*sin(1.3*x_1)*cos(x_2)$&$U(-4, 4, 100)$\\
Constant-8 & $log(x_1+1.4)+log(x1,2+1.3)$&$U(-4, 4, 100)$\\
\toprule
R1 & $\frac{(x_1+1)^3}{x_1^2-x_1+1)}$&$U(-5, 5, 100)$\\
R2 & $\frac{(x_1^2-3*x_1^2+1}{x_1^2+1)}$&$U(-4, 4, 100)$\\
R3 & $\frac{x_1^6+x_1^5)}{(x_1^4+x_1^3+x_1^2+x1+1)}$&$U(-4, 4, 100)$\\
\toprule
\newline
\end{tabular}
\end{scriptsize}
\caption{
Specific formula form and value range of the three data sets Vladislavleva and others. }
\label{a-tab3}
\end{table*}

\section{Appendix: MetSymNet tests on AIFeynman dataset.}
\label{AK}
We evaluated our proposed symbol regression algorithm, MetSymNet, using the AI Feynman dataset, which encompasses problems from physics and mathematics across various subfields like mechanics, thermodynamics, and electromagnetism. While the dataset originally included 100,000 sampled data points, we intentionally limited our analysis to 100 data points to rigorously assess MetSymNet's performance. Applying MetSymNet to perform symbol regression on each of these data points, we subsequently measured the $R^2$ values between the predicted outcomes and the correct answers.
Our experimental findings clearly demonstrate that MetSymNet adeptly captures the underlying expressions from a limited number of sample points. Remarkably, the $R^2$ values exceeded 0.99 for the majority of formulas, indicating the algorithm's proficiency in accurately fitting these expressions. These results firmly establish MetSymNet as a high-performing solution for challenges within the realms of physics and mathematics. The implications are significant, suggesting MetSymNet's potential for versatile application across various domains. You can find the detailed experimental outcomes in Table \ref{a-tab5} and Table \ref{a-tab6}.

\begin{table}[htbp]
\centering
{\footnotesize
\begin{tabular}{|l|l|r|}
\hline
Feynman   & Equation & $R^2$ \\
\hline                            
I.6.20a       & $f = e^{-\theta^2/2}/\sqrt{2\pi}$ & 0.9999  \\
I.6.20        & $f = e^{-\frac{\theta^2}{2\sigma^2}}/\sqrt{2\pi\sigma^2}$ & 0.9991\\
I.6.20b       & $f = e^{-\frac{(\theta-\theta_1)^2}{2\sigma^2}}/\sqrt{2\pi\sigma^2}$ & 0.9881 \\
I.8.14       & $d = \sqrt{(x_2-x_1)^2+(y_2-y_1)^2}$ & 0.9024  \\
I.9.18       & $F = \frac{Gm_1m_2}{(x_2-x_1)^2+(y_2-y_1)^2+(z_2-z_1)^2}$  & 0.9926\\
I.10.7       & $F = \frac{Gm_1m_2}{(x_2-x_1)^2+(y_2-y_1)^2+(z_2-z_1)^2}$  & 0.9872\\
I.11.19      & $A = x_1y_1+x_2y_2+x_3y_3$ & 0.9999   \\
I.12.1       & $F = \mu N_n$ & 1.0 \\
I.12.2       & $F = \frac{q_1q_2}{4\pi\epsilon r^2}$   & 1.0 \\
I.12.4       & $E_f = \frac{q_1}{4\pi\epsilon r^2}$  & 0.9999 \\
I.12.5       & $F = q_2 E_f$ & 1.0  \\
I.12.11      & $F = \mathcal{Q}(E_f+B v \sin\theta)$  & 0.9999 \\
I.13.4      & $K = \frac{1}{2}m(v^2+u^2+w^2)$  & 0.9982  \\
I.13.12      & $U = Gm_1m_2(\frac{1}{r_2}-\frac{1}{r_1})$ & 1.0  \\
I.14.3       & $U = mgz$ &1.0    \\
I.14.4       & $U = \frac{k_{spring}x^2}{2}$  & 0.9839  \\
I.15.3x      & $x_1 = \frac{x-ut}{\sqrt{1-u^2/c^2}}$ & 0.9793 \\
I.15.3t      & $t_1 = \frac{t-ux/c^2}{\sqrt{1-u^2/c^2}}$ & 0.9638  \\
I.15.10       & $p = \frac{m_0v}{\sqrt{1-v^2/c^2}}$ & 0.9919 \\
I.16.6       & $v_1 = \frac{u+v}{1+uv/c^2}$ & 0.9873  \\
I.18.4       & $r = \frac{m_1r_1+m_2r_2}{m_1+m_2}$ & 0.9794 \\
I.18.12      & $\tau = rF\sin\theta$  & 0.9999  \\
I.18.16      & $L = mrv \sin\theta$  & 0.9999 \\
I.24.6 & $E = \frac{1}{4} m (\omega^2+\omega_0^2) x^2$      & 0.9986\\
I.25.13      & $V_e = \frac{q}{C}$ & 1.0 \\
I.26.2       & $\theta_1 = \arcsin(n  \sin\theta_2)$ & 0.9999 \\
I.27.6       & $f_f$    $ = \frac{1}{\frac{1}{d_1}+\frac{n}{d_2}}$  & 0.9895 \\
I.29.4       & $k = \frac{\omega}{c}$ & 1.0 \\
I.29.16      & $x = \sqrt{x_1^2+x_2^2-2x_1x_2\cos(\theta_1-\theta_2)}$ & 0.9828  \\
I.30.3 & $I_* = I_{*_0}\frac{\sin^2(n\theta/2)}{\sin^2(\theta/2)}$ & 0.9912 \\
I.30.5       & $\theta = \arcsin(\frac{\lambda}{nd})$  & 0.9822\\
I.32.5       & $P = \frac{q^2a^2}{6\pi\epsilon c^3}$       & 0.9932 \\
I.32.17 & $P = (\frac{1}{2}\epsilon c E_f^2)(8\pi r^2/3) (\omega^4/(\omega^2-\omega_0^2)^2)$      & 0.9817  \\
I.34.8       & $\omega = \frac{qvB}{p}$   & 1.0\\
I.34.10       & $\omega = \frac{\omega_0}{1-v/c}$ & 0.9917 \\
I.34.14      & $\omega = \frac{1+v/c}{\sqrt{1-v^2/c^2}}\omega_0$  & 0.9992 \\
I.34.27      & $E = \hbar\omega$  & 0.9999 \\
I.37.4       & $I_* = I_1+I_2+2\sqrt{I_1I_2}\cos\delta$ & 0.9827\\
I.38.12      & $r = \frac{4\pi\epsilon\hbar^2}{mq^2}$   & 0.9999  \\
I.39.10       & $E = \frac{3}{2}p_F V$     & 0.9999 \\
I.39.11      & $E = \frac{1}{\gamma-1}p_F V$  & 0.9914 \\
I.39.22      & $P_F = \frac{n k_b T}{V}$       & 0.9976  \\
I.40.1       & $n = n_0e^{-\frac{mgx}{k_bT}}$    & 0.9816 \\
I.41.16      & $L_{rad} = \frac{\hbar\omega^3}{\pi^2c^2(e^{\frac{\hbar\omega}{k_bT}}-1)}$ & 0.9213  \\
I.43.16      & $v = \frac{\mu_{drift}q V_e}{d}$   & 0.9981  \\
I.43.31      & $D = \mu_e k_bT$    & 1.0  \\
I.43.43      & $\kappa = \frac{1}{\gamma-1}\frac{k_bv}{A}$  & 0.9215  \\
I.44.4       & $E = n k_b T \ln(\frac{V_2}{V_1})$   & 0.8017  \\
I.47.23      & $c = \sqrt{\frac{\gamma pr}{\rho}}$   & 0.9733\\
I.48.20       & $E = \frac{m c^2}{\sqrt{1-v^2/c^2}}$ &  0.8629\\
I.50.26 & $x = x_1[\cos(\omega t)+\alpha\> cos(\omega t)^2]$      & 0.9999   \\
\hline
\end{tabular}
\caption{Tested Feynman Equations, part 1.}
\label{a-tab5}
}
\end{table}
\begin{table*}[htbp]
\centering
{\footnotesize

\begin{tabular}{|l|l|r|}
\hline
Feynman   & Equation & $R^2$\\
\hline       
II.2.42   & P     $ = \frac{\kappa(T_2-T_1)A}{d}$  & 0.8015  \\
II.3.24   & $F_E = \frac{P}{4\pi r^2}$  & 0.9813 \\
II.4.23   & $V_e = \frac{q}{4\pi\epsilon r}$   & 0.9972 \\
II.6.11 & $V_e =\frac{1}{4\pi\epsilon}\frac{p_d\cos \theta}{r^2}$      & 0.9883 \\
II.6.15a & $E_f = \frac{3}{4\pi\epsilon}\frac{p_d z}{r^5} \sqrt{x^2+y^2}$      & 0.9221  \\
II.6.15b & $E_f = \frac{3}{4\pi\epsilon}\frac{p_d}{r^3} \cos\theta\sin\theta$      & 0.9917  \\
II.8.7    & $E = \frac{3}{5}\frac{q^2}{4\pi\epsilon d}$  & 0.9816  \\
II.8.31   & $E_{den} = \frac{\epsilon E_f^2}{2}$                     & 1.0 \\
II.10.9   & $E_f = \frac{\sigma_{den}}{\epsilon}\frac{1}{1+\chi}$      & 0.9999  \\
II.11.3 & $x = \frac{q E_f}{m(\omega_0^2-\omega^2)}$      & 0.9824     \\
II.11.7 & $n = n_0(1+ \frac{p_d E_f \cos\theta}{k_b T})$      & 0.8729 \\
II.11.20  & $P_* = \frac{n_\rho p_d^2 E_f}{3 k_b T}$ & 0.7225  \\
II.11.27 & $P_* = \frac{n\alpha}{1-n\alpha/3}\epsilon E_f$      & 0.9817   \\
II.11.28  & $\theta = 1+\frac{n\alpha}{1-(n\alpha/3)}$    & 0.9991\\ 
II.13.17  & $B = \frac{1}{4 \pi \epsilon c^2}\frac{2I}{r}$ & 0.9961\\
II.13.23  & $\rho_c = \frac{\rho_{c_0}}{\sqrt{1-v^2/c^2}}$          & 0.9622  \\
II.13.34  & $j = \frac{\rho_{c_0}v}{\sqrt{1-v^2/c^2}}$     & 0.9847 \\
II.15.4   & $E = -\mu_M B \cos\theta$               & 0.9999 \\
II.15.5   & $E = -p_d E_f\cos\theta$  & 0.9999 \\
II.21.32  & $V_e = \frac{q}{4\pi\epsilon r(1-v/c)}$   & 0.9915   \\
II.24.17 & $k = \sqrt{\frac{\omega^2}{c^2}-\frac{\pi^2}{d^2}}$      & 0.9872   \\
II.27.16  & $F_E = \epsilon c E_f^2$        & 0.9917 \\
II.27.18  & $E_{den} = \epsilon E_f^2$         & 0.9993 \\
II.34.2a  & $I = \frac{qv}{2\pi r}$         & 0.9916 \\
II.34.2   & $\mu_M = \frac{q v r}{2}$             & 0.9862 \\
II.34.11  & $\omega = \frac{g_{\_} q B}{2m}$          & 0.9926 \\
II.34.29a & $\mu_M = \frac{q h}{4\pi m}$      & 0.9987  \\
II.34.29b & $E = \frac{g_{\_} \mu_M B J_z}{\hbar}$ & 0.8219\\
II.35.18 & $n = \frac{n_0}{\exp(\mu_m B/(k_b T))+\exp(-\mu_m B/(k_b T))}$      & 0.9512 \\
II.35.21  & $M = n_\rho \mu_M \tanh(\frac{\mu_M B}{k_b T})$     & 0.8199 \\
II.36.38 & $f = \frac{\mu_m B}{k_b T}+\frac{\mu_m\alpha M}{\epsilon c^2 k_b T}$      & 0.9250\\
II.37.1   & $E = \mu_M(1+\chi)B$    & 0.9999\\
II.38.3   & $F = \frac{Y A x}{d}$            & 0.9999 \\
II.38.14  & $\mu_S = \frac{Y}{2(1+\sigma)}$     & 0.9999  \\
III.4.32  & $n = \frac{1}{e^{\frac{\hbar\omega}{k_bT}}-1}$ & 0.9628  \\
III.4.33  & $E = \frac{\hbar\omega}{e^{\frac{\hbar\omega}{k_b T}}-1}$  & 0.9973    \\
III.7.38  & $\omega = \frac{2 \mu_M B}{\hbar}$  & 0.9826  \\
III.8.54  & $p_{\gamma}$    $ = \sin(\frac{E t}{\hbar})^2$  & 0.9822\\
III.9.52  & $p_{\gamma}$    $ = \frac{p_d E_f t}{\hbar} \frac{    \sin((\omega-\omega_0)t/2)^2}{((\omega-\omega_0)t/2)^2}$ & 0.7024  \\
III.10.19 & $E = \mu_M\sqrt{B_x^2+B_y^2+B_z^2}$  & 0.9948 \\
III.12.43 & $L = n\hbar$ & 0.9924  \\
III.13.18 & $v = \frac{2 E d^2 k}{\hbar}$ & 0.9999  \\
III.14.14 & $I = I_0 (e^{\frac{q V_e}{k_b T}}-1)$  & 0.9914\\
III.15.12 & $E = 2U(1-\cos(kd))$    & 0.9999 \\
III.15.14 & $m = \frac{\hbar^2}{2E d^2}$     & 0.9995  \\
III.15.27 & $k = \frac{2\pi\alpha}{nd}$    & 0.9914 \\
III.17.37 & $f = \beta(1+\alpha \cos\theta)$ & 0.9988 \\
III.19.51 & $E = \frac{-mq^4}{2(4\pi\epsilon)^2\hbar^2}\frac{1}{n^2}$     & 0.9894 \\
III.21.20 & $j = \frac{-\rho_{c_0} q A_{vec}}{m}$  & 0.7262  \\
\hline
\end{tabular} 
\caption{Tested Feynman Equations, part 2.}
\label{a-tab6}
}
\end{table*}

\section{Appendix:  Computing resources} 
\label{AL}
The server we employ features an Intel(R) Xeon(R) Gold 5218R CPU, boasting a base frequency of 2.10 GHz. With a generous 20 CPU cores at its disposal, it enables seamless parallel processing, leading to enhanced computational performance. Thanks to its remarkable core count and optimized architecture, the Intel Xeon Gold 5218R proves to be exceptionally well-suited for managing resource-intensive computational tasks and workloads.

\end{document}